\documentclass[10pt,twocolumn,letterpaper]{article}

\usepackage{cvpr}
\usepackage{times}
\usepackage{epsfig}
\usepackage{graphicx}
\usepackage{amsmath}
\usepackage{amssymb}

\usepackage{colortbl}
\usepackage{float}

\definecolor{Gray}{gray}{0.85}
\definecolor{LightCyan}{rgb}{0.88,1,1}
\definecolor{LightPink}{rgb}{1,0.88,1}
\newcolumntype{G}{>{\columncolor{LightPink}}c}
\newcolumntype{B}{>{\columncolor{LightCyan}}c}

\usepackage{xcolor}

\makeatletter
\renewcommand{\paragraph}{%
  \@startsection{paragraph}{4}%
  {\z@}{1.25ex \@plus 0.8ex \@minus .4ex}{-1em}%
  {\normalfont\normalsize\bfseries}%
}
\makeatother

\usepackage{caption}
\graphicspath{{./supplement/},{./rebuttal/}}

\usepackage{subcaption}
\usepackage{multirow}
\usepackage{graphbox} 
\usepackage{booktabs} 
\usepackage{tabularx} 

\newcolumntype{Y}{>{\centering\arraybackslash}X}
\newcolumntype{C}[1]{>{\centering\let\newline\\\arraybackslash\hspace{0pt}}m{#1}}

\newcommand{\bx}{\mathbf{x}}
\newcommand{\by}{\mathbf{y}}
\newcommand{\bz}{\mathbf{z}}

\newcommand{\bs}{\mathbf{s}}

\newcommand{\bL}{\mathbf{L}}
\newcommand{\bM}{\mathbf{M}}
\newcommand{\bI}{\mathbf{I}}

\newcommand{\bmu}{\boldsymbol{\mu}}
\newcommand{\bpsi}{\boldsymbol{\psi}}
\newcommand{\btheta}{\boldsymbol{\theta}}

\newcommand{\bSigma}{\mathbf{\Sigma}}
\newcommand{\bsigma}{\boldsymbol{\sigma}}
\newcommand{\bepsilon}{\boldsymbol{\epsilon}}

\newcommand{\bzlr}{( \bz )}

\newcommand{\bmuz}{\bmu \bzlr}
\newcommand{\bSigmaz}{\bSigma_{\bpsi}\bzlr}
\newcommand{\bepsilonz}{\bepsilon \bzlr}
\newcommand{\bsigmaz}{\bsigma \bzlr}

\newcommand{\transpose}{^{\mathtt{T}}}

\newcommand{\NormalDistrib}[1]{\mathcal{N}\big(\, #1 \,\big)}

\newcommand{\bhSigma}{\hat{\bSigma}}
\newcommand{\bLambda}{\mathbf{\Lambda}}

\DeclareMathOperator*{\argmin}{arg\,min}

\usepackage[pagebackref=true,breaklinks=true,letterpaper=true,colorlinks,bookmarks=false]{hyperref}

\cvprfinalcopy 

\def\cvprPaperID{3702} 

\ifcvprfinal\fi
\begin{document}

\title{Structured Uncertainty Prediction Networks}

\author{Garoe Dorta$^{1,2}$ \quad Sara Vicente$^{2}$ \quad Lourdes Agapito$^{3}$ \quad Neill D.F. Campbell$^{1}$ \quad Ivor Simpson$^{2}$\\
$^{1}$University of Bath \qquad $^{2}$Anthropics Technology Ltd. \qquad $^{3}$University College London \\
{\tt\small $^{1}$\{g.dorta.perez,n.campbell\}@bath.ac.uk $^{2}$\{sara,ivor\}@anthropics.com $^{3}$l.agapito@cs.ucl.ac.uk}
}

\maketitle

\begin{minipage}{\textwidth}
\vspace{-35em}
Published as a conference paper at CVPR 2018
\smallskip
\hrule
\smallskip
\end{minipage}

\begin{abstract}

This paper is the first work to propose a network to predict a structured uncertainty distribution for a synthesized image.
Previous approaches have been mostly limited to predicting diagonal covariance matrices~\cite{Kingma2014VAE}.
Our novel model learns to predict a full Gaussian covariance matrix for each reconstruction, which permits efficient sampling and likelihood evaluation.

We demonstrate that our model can accurately reconstruct ground truth correlated residual distributions for synthetic datasets and generate plausible high frequency samples for real face images.  We also illustrate the use of these predicted covariances for structure preserving image denoising.
\end{abstract}


\section{Introduction}
\label{sec:introduction}

Deep probabilistic generative models have recently become the most popular tool to synthesize novel data and reconstruct unseen examples from target image distributions.
At their heart lies a density estimation problem, which for reconstruction models is commonly solved using a factorized Gaussian likelihood.
While attractive for its simplicity, this factorized Gaussian assumption comes at the cost of overly-smoothed predictions, as shown in Fig.~\ref{subfig:teaser_recons}.
In contrast, this paper introduces the first attempt to train a deep neural network to predict full structured covariance matrices to model the residual distributions of unseen image reconstructions.
We postulate that the residuals are highly structured and reflect limitations in model capacity -- we therefore propose to estimate the reconstruction uncertainty using a structured Gaussian model, to
capture pixel-wise correlations which will in turn improve the sampled reconstructions,
as shown in Fig.~\ref{subfig:teaser_covar_residual} and~\ref{subfig:teaser_covar_recons}.

\def\plotw{0.32}

\def\1img{7}

\begin{figure}[h!]
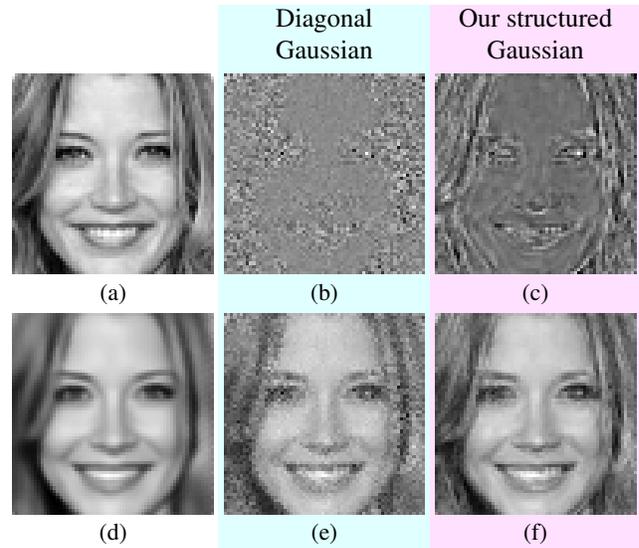

	\centering
	\captionsetup[subfigure]{aboveskip=2pt,belowskip=2pt}
	\setlength{\tabcolsep}{2pt} 
	\begin{tabular}{cBG}
	& Diagonal & Our structured \\
	& Gaussian & Gaussian \\[2pt]
	\begin{subfigure}{\plotw\linewidth}
		\includegraphics[width=\linewidth]{img/celeba/teaser_fig/input/\1img}
		\caption{}
		\label{subfig:teaser_input}
	\end{subfigure}
	&
	\begin{subfigure}{\plotw\linewidth}
		\includegraphics[width=\linewidth]{img/celeba/teaser_fig/diag/best_sample/1/\1img}
		\caption{}
		\label{subfig:teaser_diag_residual}
	\end{subfigure}
	&
	\begin{subfigure}{\plotw\linewidth}
		\includegraphics[width=\linewidth]{img/celeba/teaser_fig/covar/best_sample/1/\1img}
		\caption{}
		\label{subfig:teaser_covar_residual}
	\end{subfigure}
	\\
	\begin{subfigure}{\plotw\linewidth}
		\includegraphics[width=\linewidth]{img/celeba/teaser_fig/diag/recons/\1img} 
		\caption{}
		\label{subfig:teaser_recons}
	\end{subfigure}
	&
	\begin{subfigure}{\plotw\linewidth}
		\includegraphics[width=\linewidth]{img/celeba/teaser_fig/diag/best_recons/1/\1img} 
		\caption{}
		\label{subfig:teaser_diag_recons}
	\end{subfigure}
	&
	\begin{subfigure}{\plotw\linewidth}
		\includegraphics[width=\linewidth]{img/celeba/teaser_fig/covar/best_recons/1/\1img}
		\caption{}
		\label{subfig:teaser_covar_recons}
	\end{subfigure}
	\end{tabular}
	\caption{Given an input image~(a), a simple diagonal Gaussian likelihood model learns a smooth reconstruction~(d), with an unstructured residual sample~(b). When~(b) is added to~(d) it generates the unrealistic image~(e), demonstrating a failure to capture residual structure.
	In contrast, we learn a structured residual model, with residual samples like~(c) that, when added to (d), generate a plausible and realistic image~(f).}
	\label{fig:teaser}
\end{figure}

Learning in generative models involves some form of density estimation, where the model parameters are fitted to generate samples from the target image distribution.
This paper concentrates on methods, such as the Variational Autoencoder (VAE)~\cite{Kingma2014VAE, Rezende2014VAE_DGLM}, where an explicit representation of the data density is given by the model, however the results could in principle be extended to implicit approaches in future work.
In VAE models, a mapping is learned from a latent representation to image space.
These models commonly impose some conditions on the latent space, in the form of a prior, and the nature of the residual distribution.

It is common practice to use a Gaussian likelihood, as this provides a simple formulation for maximum likelihood estimation.
However, it is well known that samples from factorized Gaussian distributions tend to be overly-smooth. 
This effect is pronounced in common simplifications of the likelihood, such as the mean squared error, which assumes the errors at all pixels are i.i.d (independent and identically distributed) or a diagonal covariance~\cite{Kingma2014VAE,Burda2016IWAE}, which allows some local estimation of noise level but maintains the strong and flawed assumption that pixels in the residual are uncorrelated, as shown in Fig.~\ref{subfig:teaser_diag_residual} and ~\ref{subfig:teaser_covar_residual}.
These common choices of likelihood mean that if one were to draw a sample from any of these models including the noise term, white noise would be added to the reconstruction, which is unlikely to ever appear realistic, as shown in Fig.~\ref{subfig:teaser_diag_recons}.
This emphasizes the incoherence of these simplifications, and this is addressed in this paper.

This work proposes to overcome the problems of factorized likelihoods by training
a deep neural network to predict a more complex residual distribution for samples drawn from a trained probabilistic generative model.
Specifically, we model the residual distribution as a structured Gaussian distribution.
This work demonstrates that a network to predict this distribution can be tractably learned through maximum likelihood estimation of residual images; these predictions generalize well to previously unseen examples.
Samples from this model are plausible, coherent, and can capture high-frequency details that may have been missed when training the original model with a factorized likelihood.
We demonstrate the efficacy of this approach by: estimating ground truth covariances from synthetic data, adding coherent residual samples to face reconstructions from a VAE and include a further motivating example application for denoising face images.

\section{Related work}
\label{sec:related_work}
Statistical quantification of uncertainty has been an area of interest for a long time. Many traditional statistical estimation models provide some measure of uncertainty on the inferred parameters of a fitted model. An additional source of uncertainty arises from the distribution of the model residual, which is often modeled as a spherical or diagonal Gaussian distribution. 
A common issue with these models is the false assumption of independence between pixels in the reconstruction residual image. Previous work on modeling correlated Gaussian noise is limited, it has been used for small data scenarios \cite{nikias1988}, for temporally correlated noise models~\cite{WOOLRICH2001} and in Gaussian processes~\cite{rasmussen2006}.

The most recent related work on uncertainty prediction for deep generative models has been the prediction of heteroscedastic Gaussian noise for encoder/decoder models~\cite{kendall2017}.
This approach is similar to the variational autoencoder with a diagonal covariance likelihood~\cite{Kingma2014VAE}, but can be applied when the input and generated output are different, for example in semantic segmentation tasks.
Interestingly, the maps of predicted variance in~\cite{kendall2017} correspond well to high frequency structured image features, similarly to the VAE with a diagonal noise model (see Fig.~\ref{fig:celeba_variance_maps}).
These are structured regions that the model consistently struggles to accurately predict, which is a further encouragement for our work.

This paper proposes an approach to predict a structured uncertainty distribution for generated images from trained networks. This method is applicable to models that reconstruct images without embellishing the prediction with details that were not present in the original input.

Generative adversarial networks (GANs)~\cite{Goodfellow2014GAN} are an implicit density estimation method widely used for generating novel images with a great deal of success. Samples from GANs have been shown to contain fine details and can be created at very high resolutions~\cite{Karras2017progressive}.
Although, they are not designed to provide reconstructions, some methods have been proposed that enable this~\cite{Larsen2015VAEGAN, Pu2017Adversarial, Donahue2016BIGAN}.
Reconstructions from GAN models contain realistic high frequency details, however the generated images usually do not resemble well the inputs.
This might be caused by mode-dropping~\cite{Goodfellow2016Book}, where parts of the image distribution are not well modeled. Despite recent work~\cite{Arjovsky2017WGAN} addressing this issue, it remains an open problem.
Furthermore, the residual distribution for reconstruction using GAN models is likely to be highly complex and therefore we do not use a GAN model as a starting point in this work.

Explicit density methods for learning generative models use either a tractable~\cite{Oord2016PixelRNN} or approximate\cite{Rezende2014VAE_DGLM} density to model the image distribution. These models allow maximization of the likelihood of the set of training observations directly through reconstruction. These approaches generate data that is more appropriate for the task this paper addresses and offer potential likelihood models to learn the uncertainty of prediction.

To train a network to predict the uncertainty of a reconstruction model, we need to choose a reconstruction likelihood that is efficient to calculate and allows structured prediction.
PixelCNN~\cite{Oord2016PixelRNN} and derived work~\cite{Gulrajani2017PixelVAE} provide an autoregressive sampling model, where the likelihood of a pixel is modeled by a multinomial distribution conditioned on the previously generated pixels.
Although these models are capable of producing images with details, the generation process is computationally expensive. 
Approximate density models, such as the Variational Autoencoder (VAE)~\cite{Kingma2014VAE, Rezende2014VAE_DGLM}, use a variational approximation of the marginal likelihood of the data under a  diagonal Gaussian likelihood.
These models are very efficient to reconstruct from, and the Gaussian noise model has an efficient likelihood that can be extended to include off-diagonal terms to allow correlated noise prediction.
Therefore, the focus of this work is on explicit generative models with approximate likelihoods.

\section{Methodology}
\label{sec:methodology}
Many generative models, with a (diagonal) Gaussian likelihood, define the conditional probability of the target image, $\bx$, given some latent variable, $\bz$, as:
\begin{equation}
 p_{\btheta}(\bx \,|\, \bz) = \NormalDistrib{\bmuz, \bsigmaz^2 \, \bI}, 
 \label{eq:generative_model}
\end{equation}
where $\bx$ is the target image flattened to a column vector, $\btheta$ are the model parameters, and the mean $\bmuz$ and variance $\bsigmaz^2$ are (non-linear) functions of the latent variables. This is equivalent to the forward model: 
\begin{equation}
\bx = \bmuz + \bepsilonz,
\end{equation}
where $\bepsilonz \sim \NormalDistrib{\mathbf{0}, \bsigmaz^2 \bI}$ is commonly considered as unstructured noise inherent in the data.
However, reconstruction errors (or residuals) are often caused by data deficiencies, limitations in the model capacity and suboptimal model parameter estimation, while the residual itself is generally highly structured, as shown in Fig.~\ref{fig:teaser}.
The noise term is being used mostly to explain failures in the model rather than just additive noise in the input.
The novelty of this work is to encode coherent information about uncertainty in the reconstruction in $\bepsilonz$.

For the majority of models, $\bsigma^2$ is never used in practice when sampling reconstructions; it only adds white noise that would rarely improve the residual error, as shown in Fig.~\ref{fig:teaser}.
Although in some cases $\bsigma^2$ is inferred from the data, in many cases it is assumed that $\boldsymbol{\sigma}_i^2 = \sigma^2$,  a user defined constant, which simplifies the log likelihood from Eq.~\ref{eq:generative_model} to a sum of squared errors scaled by $1 / \sigma^2$.

In contrast, this paper extends the noise model to use a multivariate Gaussian likelihood with a full covariance matrix
\begin{equation}
 p_{\btheta}(\bx \,|\, \bz) = \NormalDistrib{\bmuz, \bSigmaz},
\end{equation}
where $\bSigmaz$ is a (non-linear) function parametrized by $\bpsi$; this is equivalent to $\bepsilonz \sim \NormalDistrib{\mathbf{0}, \bSigmaz}$. 
The covariance matrix captures the correlations between pixels to allow sampling of structured residuals as demonstrated in our experiments.

A maximum likelihood approach is used to train the covariance prediction. We optimize the log-likelihood with respect to $\bpsi$ keeping the generative model parameters $\btheta$ fixed:
\begin{equation}
\begin{split}
 \argmin_{\bpsi} ~ &\log\big( \big|\bSigmaz \big| \big) \; + \\
 &\; \big(\bx - \bmuz\big)\transpose \big(\bSigmaz\big)^{-1} \big(\bx - \bmuz\big).
\end{split}
\label{eq:loss_nll_covariance}
\end{equation}
To simplify notation, in subsequent sections $\bSigma$ and $\bmu$ are used to denote $\bSigmaz$ and $\bmuz$ respectively.

\section{Covariance estimation}
\label{sec:covariance_estimation}

A deep neural network is used to estimate the covariance matrix $\bSigma$ from the latent vector $\bz$, henceforth referred to as the covariance network.
This learning task is challenging on two fronts. It is ill-posed as no ground truth residual covariances are available for real data. Moreover, for each training example, a full covariance matrix must be estimated from a single target image.
By definition, the covariance matrix is symmetric and positive definite. Therefore, for an image $\bx$ with  $n$ pixels, the matrix contains $(n^2 - n) / 2 \,+ \, n$ unique parameters.

For any covariance estimation method there are three aspects to consider: (i) how difficult is it to sample from this covariance? (ii) how difficult is it to compute the terms in Eq.~\ref{eq:loss_nll_covariance}? and (iii) how difficult is it to impose symmetry and positive definiteness?

The need to draw samples arises from the fact that the covariance captures structured information about reconstruction uncertainty of the generative model.
This means that drawing a sample from $\NormalDistrib{\boldsymbol{0}, \bSigma}$ and adding that to the reconstructed output may produce a result that is more representative of the target image, as shown in Fig.~\ref{fig:teaser}.

Given a decomposition of the covariance matrix: $\bSigma = \bM \bM\transpose$,
samples can be drawn from $\NormalDistrib{\boldsymbol{\mu}, \bSigma}$ as $\mathbf{x} = \bM\mathbf{u} + \boldsymbol{\mu}$, where $\mathbf{u} \sim \NormalDistrib{\mathbf{0}, \bI}$ is a vector of standard Gaussian samples. 


If $\bSigma$ is the direct output of the covariance network, it needs to be inverted to calculate the negative log-likelihood in Eq.~\ref{eq:loss_nll_covariance}.
Hence, it is more practical to estimate the precision matrix $\bLambda = \bSigma^{-1}$ as this term appears directly in the log likelihood, and the log determinant term can be equivalently computed as $\log(|\bSigma|) = - \log(|\bLambda|)$.




\paragraph{Cholesky Decomposition} We represent the precision matrix via its Cholesky decomposition, $\bLambda = \bL \bL\transpose$, where $\bL$ is a lower triangular matrix, and the covariance network only explicitly estimates $\bL$.
%

Using the Cholesky decomposition, it is trivial to evaluate both terms in the negative log likelihood.
The reconstruction error is $\by\transpose \by$, where $\by = \bL\transpose(\bx - \bmu)$.
The log determinant is $\log(|\bSigma|) = - 2 \sum^n_i \log(l_{ii})$, where $l_{ii}$ is the $i^{\mathrm{th}}$ element in the diagonal of $\bL$.

Sampling from $\bSigma$ involves solving the triangular system of equations $\bL\transpose \mathbf{y} = \mathbf{u}$ with backwards substitution, which requires $O(n^2)$ operations.

By construction, the estimated precision matrix $\bLambda$ is symmetric. To ensure that it is also positive-definite it is sufficient to constrain the diagonal entries of $\bL$ to be strictly positive, \eg~by having the network estimate $\log(l_{ii})$ element-wise. 

However, estimating this matrix directly is only a feasible solution for datasets with small dimensionality, as the number of parameters to be estimated increases quadratically with the number of pixels in $\bx$.



\begin{figure}
  \centering
    \includegraphics[width=0.49\linewidth]{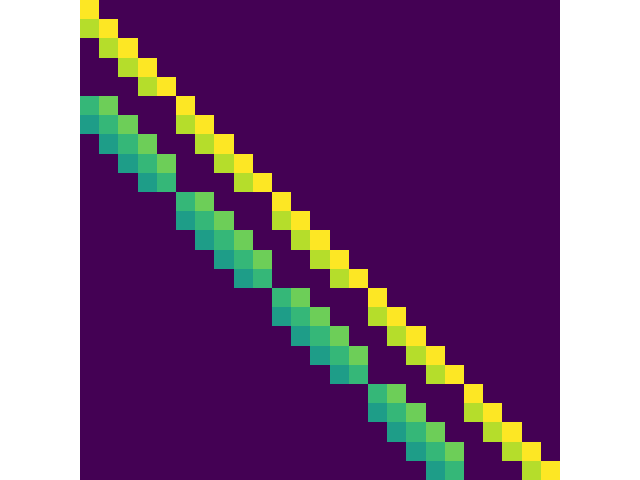}
    \includegraphics[width=0.49\linewidth]{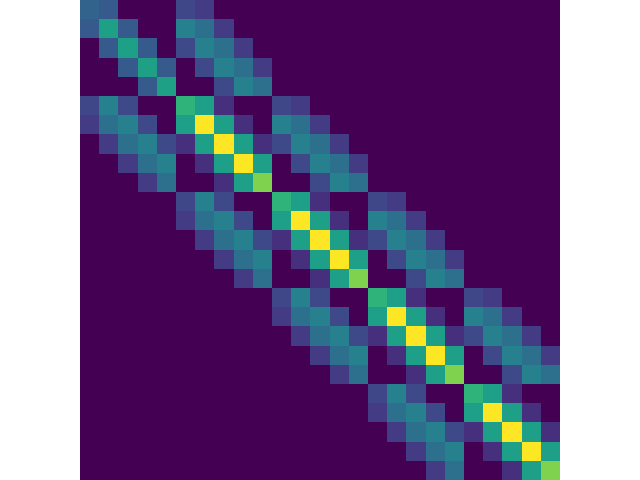}\\[2pt]
  \caption{Left, an example of the sparsity patterns in the band-diagonal lower-triangular matrices $\bL$, that are estimated by our model. Right, the precision matrix $\bLambda = \bL\bL\transpose$.}
  \label{fig:filter_sparse_covariance}
\end{figure}

\paragraph{Sparse Cholesky Decomposition} To scale to larger resolution images we can reduce complexity by imposing a fixed sparsity pattern in the matrix $\bL$, and only estimate the non-zero values of the matrix via the covariance network.

The sparsity pattern depends on the type of data being modeled.
For image data, we propose that $l_{ij}$ is only non-zero if $i\geq j$ and $i$ and $j$ are neighbours in the image plane, where pixels $i$ and $j$ are neighbours if a patch of size $f$ centred at $i$ contains $j$.
These reduces the maximum number of non-zero elements in the matrix $\bL$ to $n ((f^2 - 1) \!/  2 + 1)$, where $f << n$.

The resulting sparse matrix $\bL$ is both lower-triangular and band-diagonal as shown in Fig.~\ref{fig:filter_sparse_covariance}.
This leads to a precision matrix $\bLambda$ with a similar sparsity pattern with additional bands.

With a sparsity pattern of this form, our uncertainty model for each image can be interpreted as a Gaussian Random Field on its residual. A zero value in the precision matrix for pixels $i$ and $j$ implies that they are conditionally independent given all the other pixels. Similar Markov properties have been extensively used to model images~\cite{bishop2006pattern}.

With this representation the terms in the negative log likelihood can be evaluated efficiently, without the need to build the full dense matrix. Similarly, sampling can be performed by solving a sparse system of equations.
Moreover, this approach is amenable to parallelization on the GPU, as each patch can be evaluated independently.

\section{Results}
\label{sec:results}

We evaluate our model on two custom synthetic datasets to demonstrate the capability of our model to accurately describe known residual distributions. We also demonstrate our model on gray-scale cropped face images from the CelebA~\cite{Liu2015CelebADataset} dataset for sampling high frequency details to improve reconstructions.
Finally, we show some examples of image denoising that takes advantage of the predicted covariance to better preserve structure.
Results of our model evaluated on the CIFAR10~\cite{Krizhevsky2009Cifar10} dataset can be found in the supplemental material.
All our models are implemented in Tensorflow~\cite{Tensorflow} and they are trained on a single Titan X GPU using the Adam~\cite{Kingma2015Adam} optimizer.
Unless otherwise stated, the input data for all the experiments is normalized in $[-1,1]$.

\subsection{Synthetic datasets}
The goal of the two synthetic experiments is to evaluate the feasibility of training a covariance network to accurately estimate the residual distribution, where the true mean and covariance matrix are available for validation purposes.

Since the goal here is to evaluate the covariance prediction network, we simplify these experiments by bypassing the use of a generative model, and directly predicting $\bSigma$ from $\bmu$. 
This means that the input to our covariance prediction network is $\bmu$.

Each dataset contains 35,000 training examples and 1,000 test examples.
In all the experiments, the test examples are reconstructed by drawing a sample from the estimated covariance and adding it to the true mean $\bmu$.

Both datasets are constructed by generating a set of $\bmu$, and then adding a random sample of correlated noise to them: $\bx \sim \mathcal{N}(\bmu, \bSigma)$, where $\bSigma$ is a function of $\bmu$. Therefore, the added noise is dependent on the structure in the image, which imitates the situation on real data.
This also reflects the main assumption of our model: that there is sufficient information in the latent variables $\bz$ (or in $\bmu$ for the synthetic experiments) to estimate the residual distribution $\bSigma$.

We emphasize that despite the true covariance matrices being known for these synthetic experiments, we do not use them at train time. We train the prediction network using the objective in Eq. \ref{eq:loss_nll_covariance}, which makes no use of the true covariance and mimics the situation with real datasets.

\subsubsection{Splines}

The first synthetic dataset is composed of one dimensional signals, with 50 points per example.
Each spline is comprised of a low frequency component and a correlated high-frequency one.
The high-frequency component is produced by a unique covariance matrix per example, that is generated by a deterministic function that takes as input the low-frequency signal.
For more details please refer to the supplemental material.

The covariance prediction network is a multi-layer perceptron (MLP) with two layers of 100 units with relu activations and batch normalization~\cite{Ioffe2015BNorm}, and a final layer with 1275 units. 
The final layer directly outputs the lower triangular part of the matrix $\bL$, which does not use our proposed sparsity approach.
The model is trained with a learning rate of 1e-4 for 200 epochs. 

\def\plotw{0.49\linewidth}

\def\1img{11}
\def\2img{13}

\begin{figure}[t!]
\setlength{\tabcolsep}{0pt}
\resizebox{\linewidth}{!}{
\begin{tabular}{@{\hspace{3pt}}c @{\hspace{3pt}}c}
	\includegraphics[width=\plotw]{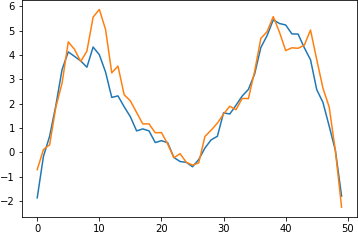} &
	\includegraphics[width=\plotw]{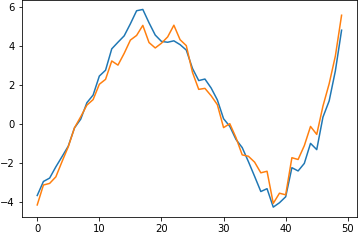} \\
	\end{tabular}
	}
	\caption{Reconstructions with samples from $\bSigma$.
	Each image corresponds to a different spline on the test dataset.
	The ground-truth spline is depicted in blue, and the estimated one in orange.}
	\label{fig:spline_reconstructions}
\end{figure}

Reconstructions for this dataset are obtained by adding a sample from $\bSigma$ to $\bmu$.
We show results for reconstructions in Fig.~\ref{fig:spline_reconstructions}, where the uncertainty model is able to add a plausible high-frequency component to the input $\bmu$.

\begin{table}[t!]
\centering\small
\setlength{\tabcolsep}{0pt}
\begin{tabularx}{\linewidth}{C{1.4cm}YYY}
	    & $-\log p(\bx \,|\, \bSigma(\bmu))$ & KL  & $||  \bSigma(\bmu) - \bSigma_{gt} ||_2$ \\ \midrule
	Ground truth & 18.65 $\pm$ 0.75 & - & - \\ \midrule 
	Diagonal model & 42.64 $\pm$ 0.83 & 72.06 $\pm$ 0.30  & 3.81 $\pm$ 1.21 \\ \midrule 
	Ours & 21.12 $\pm$ 0.79 &  3.56 $\pm$ 0.18 & 1.26 $\pm$ 0.74 \\ \midrule 
\end{tabularx} 
\caption{Quantitative comparison of reconstructions on the splines dataset.
KL denotes the KL divergence $D_{KL}( \NormalDistrib{\mathbf{0}, \bSigma(\bmu)} \,||\, \NormalDistrib{\mathbf{0}, \bSigma_{gt}} )$, where $\bSigma_{gt}$ is the ground truth covariance matrix, and  $\bSigma(\bmu)$ is the estimated covariance.
Our model obtains significant improvements under all metrics over a diagonal covariance model.
}
\label{tb:splines_error_table}
\end{table}

Quantitative results are presented in Table~\ref{tb:splines_error_table}.
We compare with a covariance network that only estimates a diagonal covariance matrix.
As the ground-truth covariances contain off-diagonal structure, the diagonal model is bound to fail in representing it.
Our model achieves a negative log likelihood similar to the one evaluated using the real covariance matrices.

\def\plotw{0.4\linewidth}

\begin{figure}[t!]
	\centering
	\begin{tabular}{cc}
	\small Ground Truth & \small Estimated \\
	\includegraphics[width=\plotw]{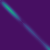} &
	\includegraphics[width=\plotw]{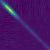} \\
	\includegraphics[width=\plotw]{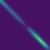} &
	\includegraphics[width=\plotw]{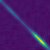} \\
	\end{tabular}
	\caption{Estimated covariance matrices for the spline dataset.
	Each row corresponds to a different spline example on the test set.
	Our model is able to learn the variations in the structured residual distributions.
	}
	\label{fig:splines_covar_estimation}
\end{figure}

As we have access to the ground-truth covariance matrices for each test example, we can directly compare the estimated covariance with the ground-truth ones. We show qualitative results in Fig.~\ref{fig:splines_covar_estimation}.
Note how the model is able to recover most of the off-diagonal values in the covariance.

\subsubsection{Ellipses}
The second synthetic dataset was built to evaluate the covariance prediction network for images and to highlight the limitations of estimating a dense matrix $\bL$.

We generate a dataset of synthetic gray-scale $16 \times 16$ images.
For each example, the mean image contains an ellipse with random width, height, position and rotation angle.
The prototype covariance matrix for this dataset generates lines and is rotated by the same random rotation angle that was used for the ellipse, thus generating random lines that are aligned with the ellipse.

For this dataset, estimating directly a dense Cholesky matrix $\bL$ requires 32,896 values per image.
Even at this limited image size we were unable to train a dense prediction model. Instead, we use the sparse prediction model with a neighborhood of size $5 \times 5$. 
The model was trained for 200 epochs with a learning rate of 1e-3, 

\def\plotw{0.23\linewidth}

\begin{figure}[t!]
	\centering
	\setlength{\tabcolsep}{1pt} 
	\begin{tabular}{ccc}
	$\bmu$ & $\bx$  & $\bmu + \bepsilon$ \\
	\includegraphics[width=\plotw]{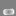} &
	\includegraphics[width=\plotw]{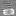} &
	\includegraphics[width=\plotw]{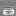} \\
	\includegraphics[width=\plotw]{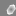} &
	\includegraphics[width=\plotw]{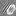} &
	\includegraphics[width=\plotw]{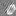} \\
	\includegraphics[width=\plotw]{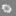} &
	\includegraphics[width=\plotw]{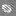} &
	\includegraphics[width=\plotw]{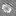} \\
	\end{tabular}
	\caption{Reconstructions from our model, left column: input $\bmu$, middle column: original $\bx$, and right column: reconstruction using a sample from the learned residual distribution, where $\bepsilon \sim \NormalDistrib{\mathbf{0}, \bSigma}$.
	}
	\label{fig:ellipses_reconstructions}
\end{figure}

Reconstructions of the test set are shown in Fig.~\ref{fig:ellipses_reconstructions}, where we show results of taking a sample from $\bSigma$ which is added to $\bmu$.
The covariance prediction network is successful in mapping the uncertainty distribution from the mean $\bmu$. The samples from the covariance exhibit high-frequency detail that matches the true residual. \begin{table}[t!]
\centering\small
\setlength{\tabcolsep}{0pt}
\begin{tabularx}{\linewidth}{C{1.2cm}YYY}
	& $-\log p(\bx \,|\, \bSigma(\bmu))$ & KL  & $||  \bSigma(\bmu) - \bSigma_{gt} ||_2$ \\ \midrule
	Ground truth & -286 $\pm$ 2.8  & -  & - \\ \midrule 
	Diagonal model & -149 $\pm$ 3.3 & 707 $\pm$ 7.2 & 1.80 $\pm$ 0.21 \\ \midrule 
	Ours & -259 $\pm$ 2.7  & 113 $\pm$ 2.6  & 1.06 $\pm$ 0.34 \\ \midrule 
\end{tabularx} 
\caption{Quantitative comparison on the ellipses dataset (see Table~\ref{tb:splines_error_table} for a description of the metrics).
Our model is able to better model the real covariance matrices with its more complex uncertainty distribution.
}
\label{tb:ellipses_error_table}
\end{table}
A quantative comparison with a diagonal Gaussian model is presented in Table~\ref{tb:ellipses_error_table}.
Our model achieves a negative log likelihood similar to the one evaluated using the real covariance matrices.

\def\splDataW{0.4}

\begin{figure}[t!]
	\centering
	\begin{tabular}{cc}
	\small Ground Truth & \small Estimated \\
	\includegraphics[width=\splDataW\linewidth]{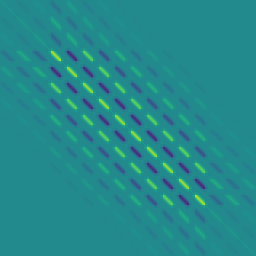} &
	\includegraphics[width=\splDataW\linewidth]{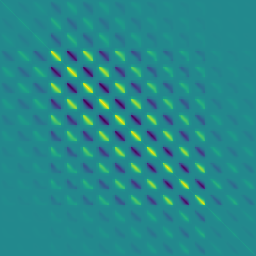} \\
	\includegraphics[width=\splDataW\linewidth]{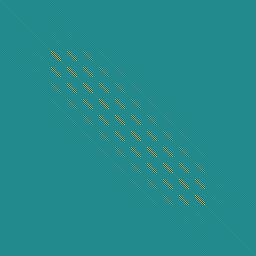} &
	\includegraphics[width=\splDataW\linewidth]{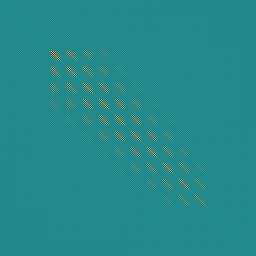} \\
	\end{tabular}
	\caption{Covariance matrices estimated by our model, where each row corresponds to a different ellipse example on the test set.	
	 Much of the structure of the real covariance matrices is recovered.}
	\label{fig:ellipses_covar_estimation}
\end{figure}

Examples of ground-truth and estimated covariances are shown in Fig.~\ref{fig:ellipses_covar_estimation} and illustrate the accuracy of the covariance prediction. The covariance structure for this model is more complex than the one for the splines, and yet the model is still able to recover it effectively with our sparse estimation of the Cholesky matrix.

\subsection{CelebA}

We now show results of employing the covariance prediction network on a real images dataset: CelebA~\cite{Liu2015CelebADataset}.
The aligned and cropped version of the dataset is used, where a further cropping and resizing to $64 \times 64$ is performed and the images are converted from RGB to gray-scale.
The dataset consists of 202,599 images of faces, which we split into 182,637 for training and 19,962 for testing as recommended by the authors.

We train both an Autoencoder (AE)~\cite{Cottrell1987AE} and a VAE on this dataset using the architecture in~\cite{Pu2017Adversarial}.
The Autoencoder is trained with a mean squared error, and it serves to show the performance of our model when using as input a $\bz$ from an uncontrolled latent space.
We then trained two covariance prediction networks, one for each model, to estimate the residual uncertainty from the latent variables, $\bz$.
These networks have an initial block that is similar to the decoder of the VAE and a second block with four convolutional layers.
The output of the networks is the sparse Cholesky decomposition of the precision matrix with a neighborhood of size $7 \times 7$ pixels.
The sparsity imposed on the matrix means that instead of estimating $8,390,656$ values as output, the network only needs to estimate $102,400$.
The covariance networks are trained with a learning rate of 1e-3 for 50 epochs.
Additional implementation details are given in the supplemental material.

\def\plotw{0.19\linewidth}

\def\1img{9}
\def\2img{11}
\def\3img{1}
\def\4img{13}

\begin{figure}[t!]
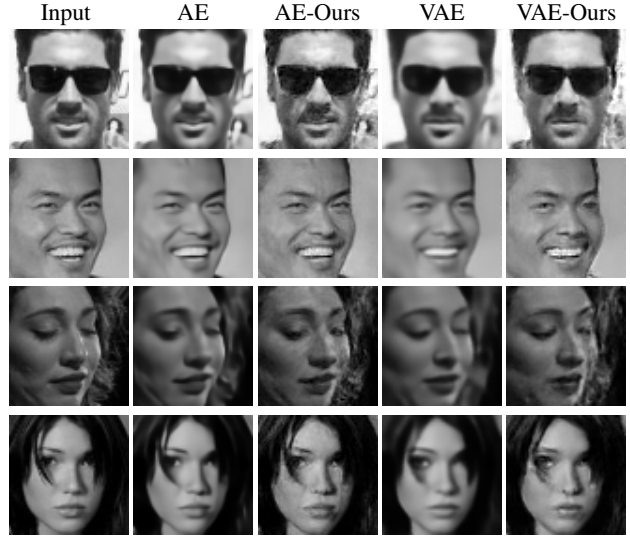

	\centering
	\setlength{\tabcolsep}{1pt} 
	\begin{tabular}{cccccc}
	{\small Input } & {\small AE} & {\small AE-Ours} & {\small VAE} & {\small VAE-Ours} & \\
	\includegraphics[width=\plotw]{img/celeba/recons_fig/input/\1img} &
	\includegraphics[width=\plotw]{img/celeba/recons_fig/ae/recons/\1img} &
	\includegraphics[width=\plotw]{img/celeba/recons_fig/ae/best_recons/1/\1img} &
	\includegraphics[width=\plotw]{img/celeba/recons_fig/vae/recons/\1img} &
	\includegraphics[width=\plotw]{img/celeba/recons_fig/vae/best_recons/1/\1img}
	\\
	\includegraphics[width=\plotw]{img/celeba/recons_fig/input/\2img} &
	\includegraphics[width=\plotw]{img/celeba/recons_fig/ae/recons/\2img} &
	\includegraphics[width=\plotw]{img/celeba/recons_fig/ae/best_recons/1/\2img} &
	\includegraphics[width=\plotw]{img/celeba/recons_fig/vae/recons/\2img} &
	\includegraphics[width=\plotw]{img/celeba/recons_fig/vae/best_recons/1/\2img}
	\\
	\includegraphics[width=\plotw]{img/celeba/recons_fig/input/\3img} &
	\includegraphics[width=\plotw]{img/celeba/recons_fig/ae/recons/\3img} &
	\includegraphics[width=\plotw]{img/celeba/recons_fig/ae/best_recons/1/\3img} &
	\includegraphics[width=\plotw]{img/celeba/recons_fig/vae/recons/\3img} &
	\includegraphics[width=\plotw]{img/celeba/recons_fig/vae/best_recons/1/\3img}
	\\
	\includegraphics[width=\plotw]{img/celeba/recons_fig/input/\4img} &
	\includegraphics[width=\plotw]{img/celeba/recons_fig/ae/recons/\4img} &
	\includegraphics[width=\plotw]{img/celeba/recons_fig/ae/best_recons/1/\4img} &
	\includegraphics[width=\plotw]{img/celeba/recons_fig/vae/recons/\4img} &
	\includegraphics[width=\plotw]{img/celeba/recons_fig/vae/best_recons/1/\4img}
	\end{tabular}
	\caption{Comparison of image reconstructions for the different models.
	The AE and VAE both generate over-smoothed images.
	For both the AE and VAE, our model adds plausible high-frequencies from a single sample drawn from the predicted uncertainty distribution.} 
	\label{fig:celeba_reconstructions}
\end{figure}

Example reconstructions for the VAE and the AE models are shown in Fig.~\ref{fig:celeba_reconstructions}.
Reconstructions from both methods suffer from the over-smoothing effects of using a simplified Gaussian likelihood and in both cases high frequency details are lost. By adding a random sample of the predicted residual distribution our method is able to recover plausible high-frequency details, resulting in more realistic looking images. The added detail corresponds to important face features, which is lost by both autoencoder models, like teeth and hair.
\begin{table}[t!]
\centering
\setlength{\tabcolsep}{2pt}
\begin{tabularx}{\linewidth}{YYYY}
	\textbf{Model}    & NLL & $-\log p(\bx \,|\, \bz)$ \\ \midrule
	AE~\cite{Cottrell1987AE} & - & - \\ \midrule
	VAE~\cite{Kingma2014VAE}  &  $-5378 \pm \phantom{0}931$  & $ -6079 \pm \phantom{0}936$ \\ \midrule
	Ours-AE & - & $-8242 \pm \phantom{0}433$   \\ \midrule	
	Ours-VAE & $-7753 \pm 1323$ & $-8386 \pm 1339$  \\ \midrule
\end{tabularx} 
\caption{Quantitative comparison of density estimation error.
NLL denotes the upper bound of the marginal negative log likelihood, lower is better.
The residuals are structured, thus the diagonal model produces poor estimations.
Our estimation method is able to significantly improve over the AE and VAE simplified noise model.
}
\label{tb:celeba_error_table}
\end{table}


A quantitative comparison of using either the predicted covariance or a diagonal covariance for calculating the negative log-likelihood of the reconstructions $\bmu$ is presented in Table~\ref{tb:celeba_error_table}.
The reported values for VAE-based methods follow the protocol described in~\cite{Burda2016IWAE}, where the upper bound of the marginal negative log likelihood is evaluated by numerically integrating over $\bz$ with 500 samples per image.
For VAE and our model using as base a VAE, the $-\log p(\bx \,|\, \bz)$ term is evaluated imposing zero variance in $\bz$.

\def\plotw{0.225\linewidth}

\def\1img{0}
\def\2img{2}
\def\3img{14}
\def\4img{41}
\def\5img{4}

\begin{figure}[t!]
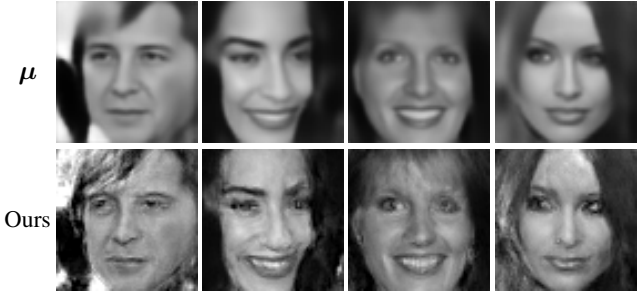

	\centering
	\setlength{\tabcolsep}{1pt} 
	\begin{tabular}{ccccc}
	{$\bmu$} & 
	\includegraphics[align=c,width=\plotw]{img/celeba/sample_cov_sample_rebuttal1/sample/\1img} &
	\includegraphics[align=c,width=\plotw]{img/celeba/sample_cov_sample_rebuttal1/sample/\2img} &
	\includegraphics[align=c,width=\plotw]{img/celeba/sample_cov_sample_rebuttal1/sample/\3img} &
	\includegraphics[align=c,width=\plotw]{img/celeba/sample_cov_sample_rebuttal1/sample/\4img} \\[23pt]
	{\small Ours} & 
	\includegraphics[align=c,width=\plotw]{img/celeba/sample_cov_sample_rebuttal1/joint_samples/0/\1img} &
	\includegraphics[align=c,width=\plotw]{img/celeba/sample_cov_sample_rebuttal1/joint_samples/0/\2img} & 
	\includegraphics[align=c,width=\plotw]{img/celeba/sample_cov_sample_rebuttal1/joint_samples/1/\3img} & 
	\includegraphics[align=c,width=\plotw]{img/celeba/sample_cov_sample_rebuttal1/joint_samples/1/\4img} \\[23pt]
	\end{tabular}
	\caption{Samples from a $\beta$-VAE with residual samples from our model.
	The covariance network predicts plausible structured residuals for the synthesized images.}
	\label{fig:celeba_samples_retrain}
\end{figure}
Images generated by decoding samples from the prior distribution on the latent space of a $\beta$-VAE with added residuals from our model are shown in Fig.~\ref{fig:celeba_samples_retrain}.
We found that the VAE with a diagonal Gaussian likelihood overfitted the reconstruction error, thus neglecting the KL term for the prior on the latent space, which in turn produces low quality samples.
Instead, we trained our structured uncertainty network on a $\beta$-VAE~\cite{Higgins2017BVAE} with $\beta = 5$.
This corresponds to increasing the weight of the KL term on a VAE, which is known to improve sample quality.
Our model is able to generate structured residuals of similar quality as those whose $\bz$ was created by encoding an image.
The covariance network is still able to learn meaningful structured uncertainty for the VAE model, as shown in the supplemental material.

\def\plotw{0.24\linewidth}
\def\ploti{3} 
\def\6img{8}

\begin{figure}[t!]
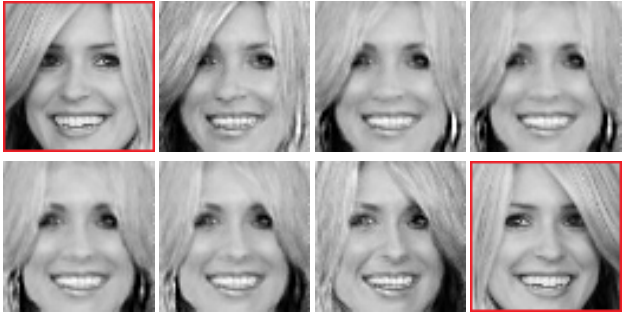

	\centering
	\setlength{\tabcolsep}{1pt} 
	\begin{tabular}{cccc}
	\includegraphics[width=\plotw]{supplement/supp_img/celeba/interp_fig/input_source/\6img/0} &
	\includegraphics[width=\plotw]{supplement/supp_img/celeba/interp_fig/recons_with/\6img/0} &
	\includegraphics[width=\plotw]{supplement/supp_img/celeba/interp_fig/recons_with/\6img/2} &
	\includegraphics[width=\plotw]{supplement/supp_img/celeba/interp_fig/recons_with/\6img/4} \\
	\includegraphics[width=\plotw]{supplement/supp_img/celeba/interp_fig/recons_with/\6img/6} &
	\includegraphics[width=\plotw]{supplement/supp_img/celeba/interp_fig/recons_with/\6img/8} & 
	\includegraphics[width=\plotw]{supplement/supp_img/celeba/interp_fig/recons_with/\6img/10} &
	\includegraphics[width=\plotw]{supplement/supp_img/celeba/interp_fig/input_target/\6img/0}
	\end{tabular}
	\caption{ Samples drawn with our model while interpolating on the latent space, from the top-left input to the bottom-right one.
	Using a fixed noise vector $\mathbf{u}$, samples are drawn from our model as $\bx = \bmu + \bM \mathbf{u}$, where the covariance $\bSigma = \bM\bM^T$.
	The residuals are smoothly interpolated for the different images, which suggests that the estimated covariance matrices also vary smoothly.}
	\label{fig:celeba_samples_interpolation}
\end{figure}

To evaluate the generalization of the model to different regions in the latent space, we show in Fig.~\ref{fig:celeba_samples_interpolation} the result of interpolating between an image and its x-flipped mirror image. 
The generated images are plausible and the sampled residuals are consistent across the interpolated images.

\def\plotw{0.12\linewidth}

\def\1img{7}
\def\2img{14}
\def\3img{13}

\begin{figure*}[t!]
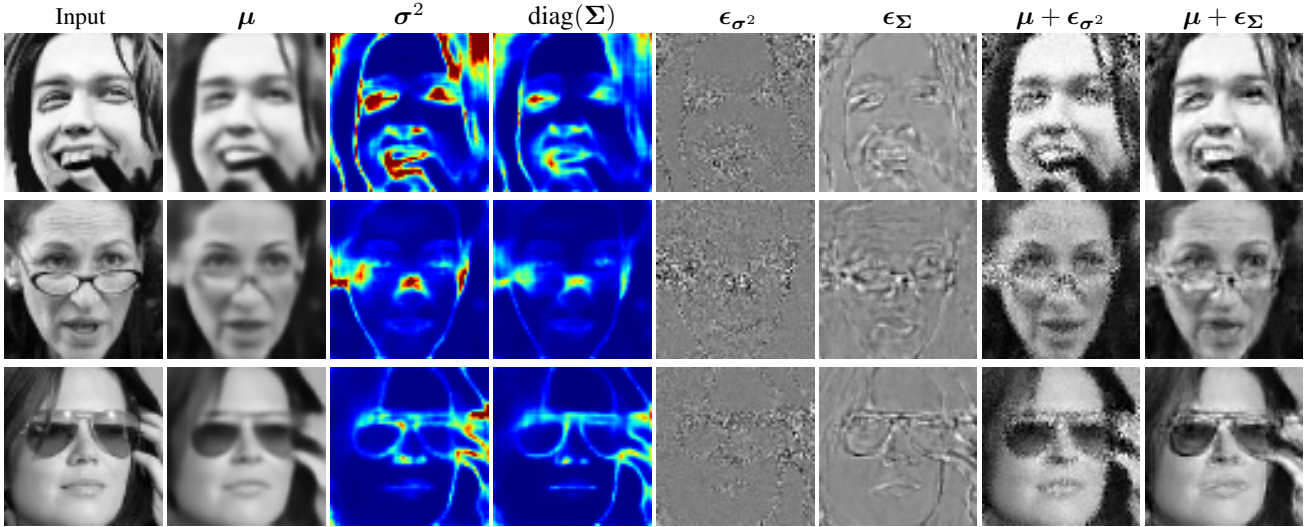

	\centering
	\setlength{\tabcolsep}{1pt} 
	\begin{tabular}{cccccccc}
	{\small Input } & $\bmu$ & $\bsigma^2$ & $\text{diag}(\bSigma)$ & { $\bepsilon_{\bsigma^2}$} & $\bepsilon_{\bSigma}$ & $\bmu + \bepsilon_{\bsigma^2}$  & $\bmu + \bepsilon_{\bSigma}$\\
	\includegraphics[width=\plotw]{img/celeba/variance_fig/input/\1img} &
	\includegraphics[width=\plotw]{img/celeba/variance_fig/recons/\1img} &
	\includegraphics[width=\plotw]{img/celeba/variance_fig/variance_diag/\1img} &
	\includegraphics[width=\plotw]{img/celeba/variance_fig/variance_covar/\1img} &
	\includegraphics[width=\plotw]{img/celeba/variance_fig/diag/residuals/\1img} &
	\includegraphics[width=\plotw]{img/celeba/variance_fig/covar/residuals//\1img} &
	\includegraphics[width=\plotw]{img/celeba/variance_fig/diag/recons_samples/\1img} &
	\includegraphics[width=\plotw]{img/celeba/variance_fig/covar/recons_samples/\1img}
	\\
	\includegraphics[width=\plotw]{img/celeba/variance_fig/input/\2img} &
	\includegraphics[width=\plotw]{img/celeba/variance_fig/recons/\2img} &
	\includegraphics[width=\plotw]{img/celeba/variance_fig/variance_diag/\2img} &
	\includegraphics[width=\plotw]{img/celeba/variance_fig/variance_covar/\2img} &
	\includegraphics[width=\plotw]{img/celeba/variance_fig/diag/residuals/\2img} &
	\includegraphics[width=\plotw]{img/celeba/variance_fig/covar/residuals//\2img} &
	\includegraphics[width=\plotw]{img/celeba/variance_fig/diag/recons_samples/\2img} &
	\includegraphics[width=\plotw]{img/celeba/variance_fig/covar/recons_samples/\2img}
	\\
	\includegraphics[width=\plotw]{img/celeba/variance_fig/input/\3img} &
	\includegraphics[width=\plotw]{img/celeba/variance_fig/recons/\3img} &
	\includegraphics[width=\plotw]{img/celeba/variance_fig/variance_diag/\3img} &
	\includegraphics[width=\plotw]{img/celeba/variance_fig/variance_covar/\3img} &
	\includegraphics[width=\plotw]{img/celeba/variance_fig/diag/residuals/\3img} &
	\includegraphics[width=\plotw]{img/celeba/variance_fig/covar/residuals//\3img} &
	\includegraphics[width=\plotw]{img/celeba/variance_fig/diag/recons_samples/\3img} &
	\includegraphics[width=\plotw]{img/celeba/variance_fig/covar/recons_samples/\3img}
	\end{tabular}
	\caption{ Variance maps for different inputs, where $\bmu$ and $\bsigma^2$ are predicted by a diagonal covariance VAE, and $\text{diag}(\bSigma)$ is the diagonal of our estimated covariance matrix.
	Residual predictions are sampled as $\bepsilon_{\bsigma^2} \sim \NormalDistrib{\mathbf{0}, \bsigma^2 \bI}$ for the VAE, and $\bepsilon_{\bSigma} \sim \NormalDistrib{\mathbf{0}, \bSigma}$ for our model.
	The diagonal noise estimation model mistakenly identifies teeth or skin wrinkles as variance, whereas the covariance model properly identifies them as regions with high covariance, yet low variance.}
	\label{fig:celeba_variance_maps}
\end{figure*}

To further highlight the differences between the diagonal and predicted covariance noise models, the variance maps for both are shown in Fig.~\ref{fig:celeba_variance_maps}.
The diagonal model must explain all the errors with variance, while our model is able to explain some with correlations. The effect of this is evident when sampling from the estimated $\bSigma$.

\subsection{Denoising Example}
\def\plotw{0.12\linewidth}

\def\1img{10}
\def\2img{5}
\def\3img{8}
\def\4img{15}
\def\nEig{1000} 

\begin{figure*}[t!]
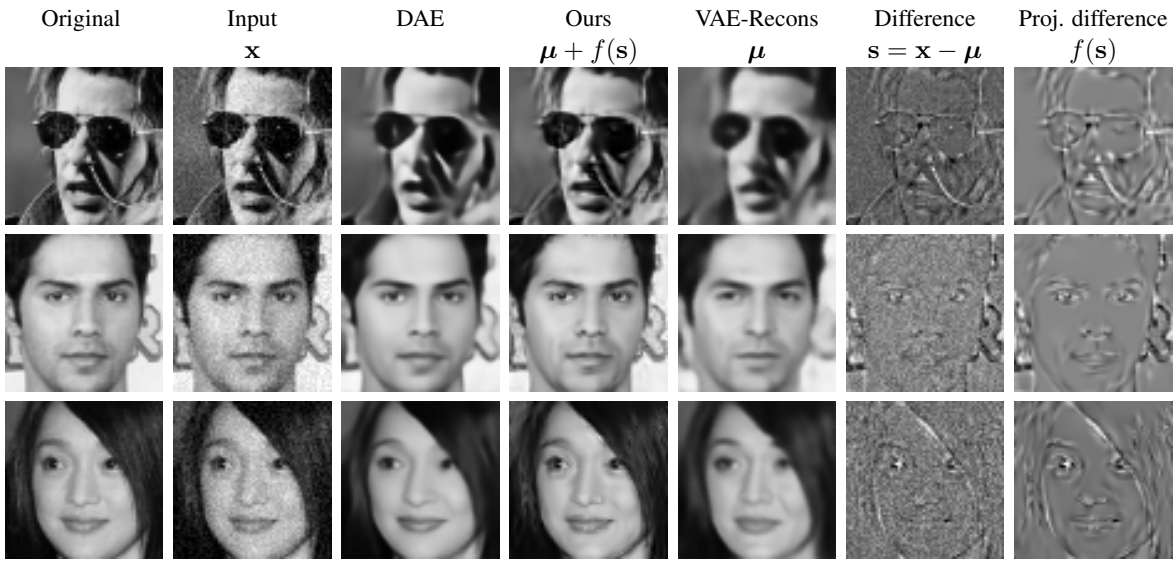

	\centering
	\setlength{\tabcolsep}{2pt} 
	\begin{tabular}{ccccccc}
	{\small Original } & {\small Input} & {\small DAE} & {\small Ours} &  {\small VAE-Recons} &
	{\small Difference} & {\small Proj. difference} \\
	& $\bx$ & & $\bmu + f(\bs)$ & $\bmu$ & $\bs = \bx - \bmu$ & $f(\bs)$ \\
	\includegraphics[width=\plotw]{img/celeba/denoising/input/\1img} &
	\includegraphics[width=\plotw]{img/celeba/denoising/input_noisy/\1img} &
	\includegraphics[width=\plotw]{img/celeba/denoising/denoising_ae/\1img} &
	\includegraphics[width=\plotw]{img/celeba/denoising/covar/denoised/\nEig/\1img} &
	\includegraphics[width=\plotw]{img/celeba/denoising/covar/recons/\1img} &
	\includegraphics[width=\plotw]{img/celeba/denoising/covar/residuals_noise/\1img} &
	\includegraphics[width=\plotw]{img/celeba/denoising/covar/proj_noise/\nEig/\1img} \\
	\includegraphics[width=\plotw]{img/celeba/denoising/input/\2img} &
	\includegraphics[width=\plotw]{img/celeba/denoising/input_noisy/\2img} &
	\includegraphics[width=\plotw]{img/celeba/denoising/denoising_ae/\2img} &
	\includegraphics[width=\plotw]{img/celeba/denoising/covar/denoised/\nEig/\2img} &
	\includegraphics[width=\plotw]{img/celeba/denoising/covar/recons/\2img} &
	\includegraphics[width=\plotw]{img/celeba/denoising/covar/residuals_noise/\2img} &
	\includegraphics[width=\plotw]{img/celeba/denoising/covar/proj_noise/\nEig/\2img} \\
	\includegraphics[width=\plotw]{img/celeba/denoising/input/\3img} &
	\includegraphics[width=\plotw]{img/celeba/denoising/input_noisy/\3img} &
	\includegraphics[width=\plotw]{img/celeba/denoising/denoising_ae/\3img} &
	\includegraphics[width=\plotw]{img/celeba/denoising/covar/denoised/\nEig/\3img} &
	\includegraphics[width=\plotw]{img/celeba/denoising/covar/recons/\3img} &
	\includegraphics[width=\plotw]{img/celeba/denoising/covar/residuals_noise/\3img} &
	\includegraphics[width=\plotw]{img/celeba/denoising/covar/proj_noise/\nEig/\3img}
	\end{tabular}
	\caption{
	 Denoising experiment, left column: original image without noise, second column: image with added noise, third column: denoising autoencoder (DAE) result, fourth column: our result, fifth column: VAE reconstruction from the noisy input, sixth column: difference between the VAE reconstruction and the noisy input, right column: the difference projected on $\hat{\bSigma}$, the matrix constructed with 1000 eigenvectors of $\bSigma$.
	Our result is the sum of the projected difference and the VAE reconstruction.
	Our model is able to recover fine details that are lost with the DAE approach.}
	\label{fig:celeba_denoising}
\end{figure*}

One possible application of the covariance prediction network is image denoising, as shown in Fig.~\ref{fig:celeba_denoising}.
We hypothesize that our predicted covariance matrices will only span the space of valid face residuals, thus projecting a noisy residual in that space will remove the noise.
Here, the noisy image is reconstructed using a VAE, that was \emph{not} trained with noisy data.
The difference between the noisy input and the reconstruction is computed, and projected onto $\bhSigma$, which is created by taking the 1000 eigenvectors of $\bSigma$ with the largest eigenvalues.
The projected difference is added to the VAE reconstruction to produce the final denoised image.
A comparison is shown with an Autoencoder trained explicitly for denosing with the same architecture as the VAE.
Note how the use of the structure covariance model is able to filter the noise to generate plausible structured high-frequency details, while the denoising Autoencoder fails to recover those details.

\begin{table}[t!]
\centering
\setlength{\tabcolsep}{2pt}
\begin{tabularx}{\linewidth}{YY}
	\textbf{Model}    & \textbf{MSE} \\ \midrule
	DAE &  5.13e-3 $\pm$ 2.52e-3  \\ \midrule  
	Ours & 2.99e-3 $\pm$ 7.98e-4   \\ \midrule	
\end{tabularx} 
\caption{Quantitative comparison for denoising in terms of mean squared error (MSE) with respect to the noise-free input.
Our model is on average able to produce better results than an Autoencoder trained  for denoising.
}
\label{tb:celeba_denoising_error_table}
\end{table}

Quantitative results for this experiment are shown in Table~\ref{tb:celeba_denoising_error_table}, where MSE is reported for the first 2000 images in the test set. 
Our model achieves significantly lower error than an Autoencoder trained specifically for this task.

%

\section{Conclusions}
\label{sec:conclusions}
In this paper, we have demonstrated what we believe is the first attempt to train a deep neural network to predict structured covariance matrices for the residual image distribution of unseen image reconstructions. Our results show that ground truth covariances can be learned for toy data and good samples generated for celebA data. A further motivating experiment is also shown for denoising face images.

An interesting question is whether we can train a generative model and covariance network in tandem to produce better reconstructions. We postulate that a better residual model would improve the reconstructions, as it may consider plausible but different realizations of high frequency image features as likely. However, there may be some effort required to keep the two networks consistent and additional investigations are needed into priors on the predicted covariances.

Another questions is what is the best input data to use for the covariance network. Is it best to learn directly from $\bz$ or $\bmu$ or some other pre-learned function of either of these?

Finally, in this work we have only examined the residual image distribution for reconstruction models trained with a Gaussian likelihood. An interesting avenue to explore would be a structured pixel uncertainty distribution for other reconstruction error metrics, such as perceptual loss. 

\newpage
{
\bibliographystyle{ieee}
\bibliography{bibliography}
}

\newpage
\appendix
\onecolumn
\section{Network architectures}
\label{sec:network_architecture}
All the models were trained with a batch size of 64.
The $\exp$ block in all the architectures removes the $\log$ in the diagonal values of the Cholesky matrix that is being estimated.

\begin{minipage}{\linewidth}
	\begin{minipage}[b]{.46\textwidth}
	\begin{figure}[H]
	\center
	\includegraphics[width=0.45\linewidth]{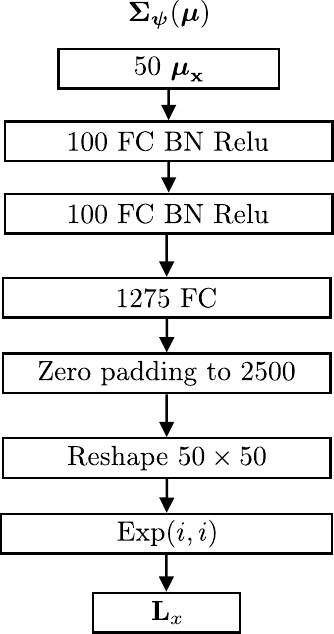}
	\caption{Covariance estimation network architecture for the splines dataset.}
	\label{fig:splines_covar_arch}
\end{figure}
	\end{minipage}
	\hfill
	\begin{minipage}[b]{.49\textwidth}
	\begin{figure}[H]
	\center
	\includegraphics[width=0.7\linewidth]{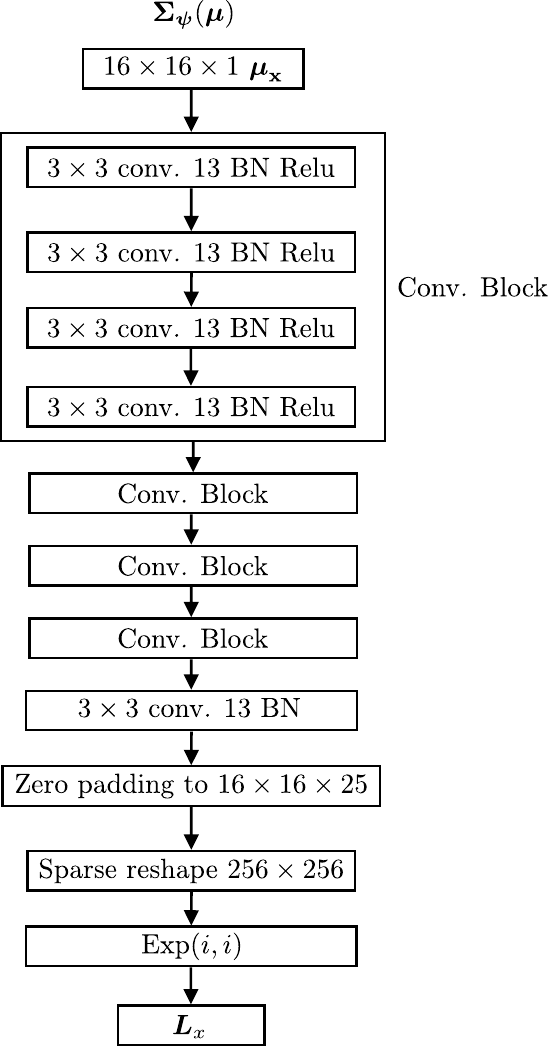}
	\caption{Covariance estimation network architecture for the ellipses dataset.}
	\label{fig:ellipses_covar_arch}
\end{figure}
	\end{minipage}
\end{minipage}


\begin{figure}[H]
	\center
	\includegraphics[width=0.9\linewidth]{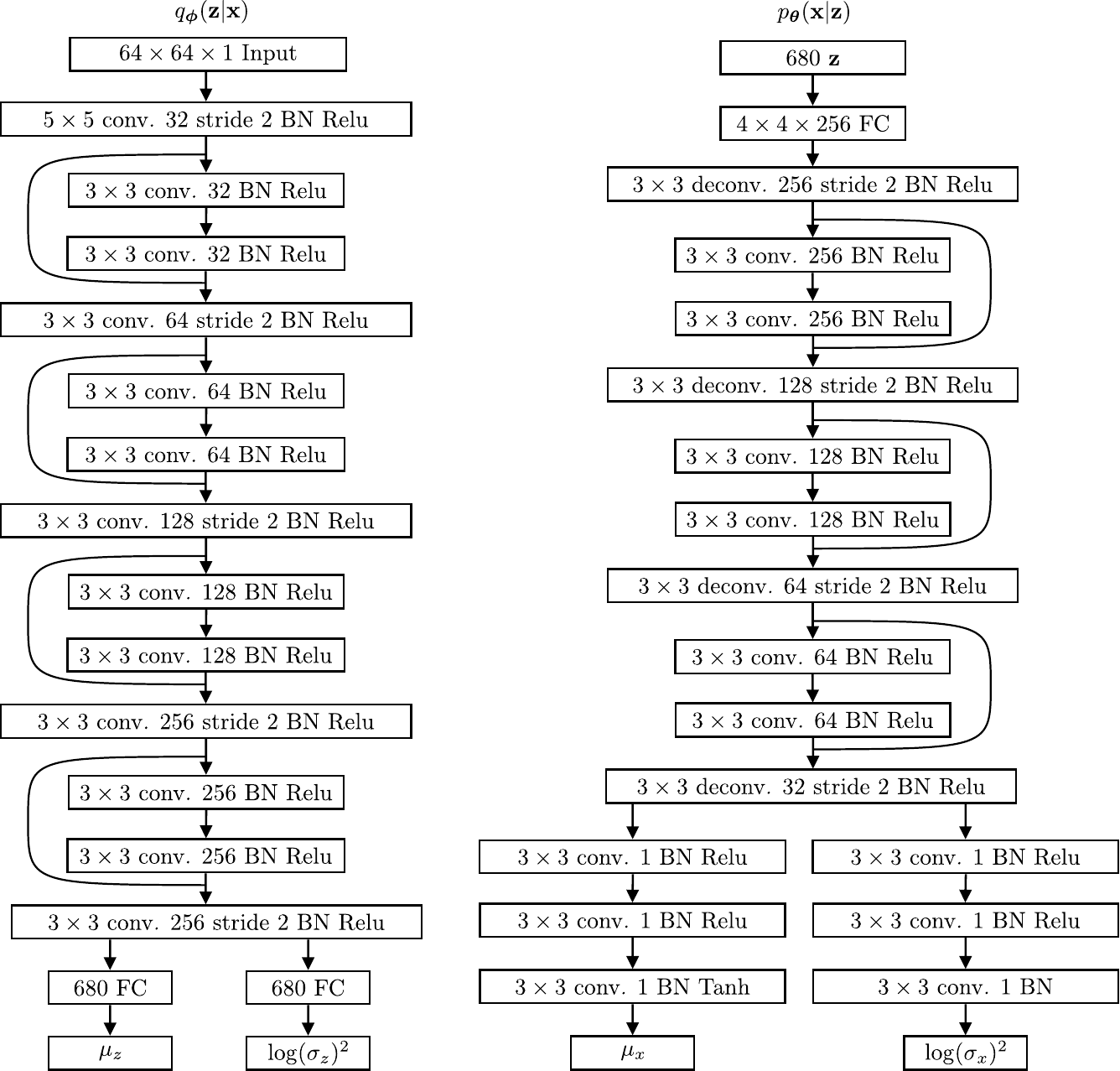}
	\caption{VAE architecture for the CelebA dataset.}
	\label{fig:celeba_vae_arch}
\end{figure}

\begin{figure}[H]
	\center
	\includegraphics[width=0.3\linewidth]{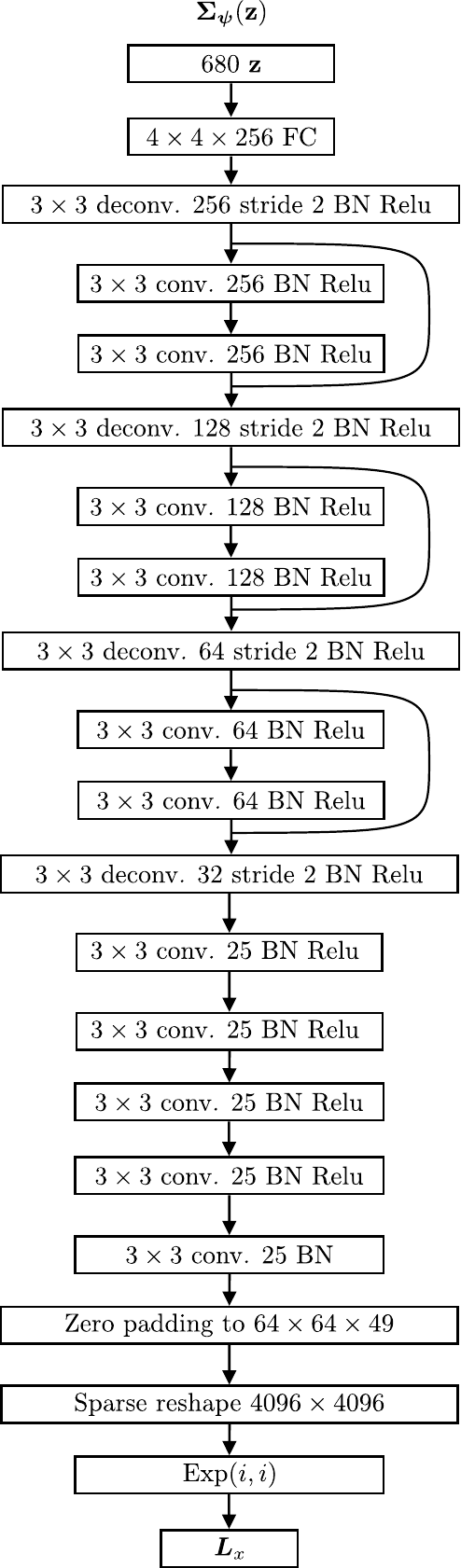}
	\caption{Covariance estimation network architecture for the CelebA dataset.}
	\label{fig:celeba_covar_arch}
\end{figure}

\newpage
\section{CIFAR10 dataset}
\label{sec:cifar10_datasets}
In this dataset our models uses a patch size of $5 \times 5$, and an architecture adapted from the CelebA model for $32 \times 32$ images.
Quantitatively, a VAE with diagonal covariance achieved a marginal negative log likelihood of  $-1026 \pm 462$ and our model of $-1333 \pm 477$, where the likelihood was evaluated with 500 $\bz$ samples per image.

\def\plotw{0.095\linewidth}

\begin{figure}[btph]
	\centering
	\setlength{\tabcolsep}{1pt} 
	\begin{tabular}{cccc  @{\hspace{5\tabcolsep}}  |  @{\hspace{5\tabcolsep}}  cccc}
	{\small Input } & {\small $\bmu$} & {\small Diag} & {\small Ours} & {\small Input } & {\small $\bmu$} & {\small Diag} & {\small Ours} \\
	\includegraphics[width=\plotw]{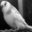} &
	\includegraphics[width=\plotw]{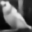} &
	\includegraphics[width=\plotw]{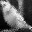} &
	\includegraphics[width=\plotw]{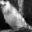} &
	\includegraphics[width=\plotw]{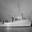} &
	\includegraphics[width=\plotw]{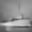} &
	\includegraphics[width=\plotw]{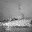} &
	\includegraphics[width=\plotw]{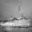} \\
	\includegraphics[width=\plotw]{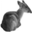} &
	\includegraphics[width=\plotw]{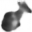} &
	\includegraphics[width=\plotw]{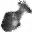} &
	\includegraphics[width=\plotw]{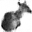} &
	\includegraphics[width=\plotw]{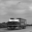} &
	\includegraphics[width=\plotw]{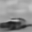} &
	\includegraphics[width=\plotw]{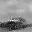} &
	\includegraphics[width=\plotw]{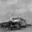} \\
	\includegraphics[width=\plotw]{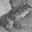} &
	\includegraphics[width=\plotw]{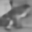} &
	\includegraphics[width=\plotw]{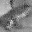} &
	\includegraphics[width=\plotw]{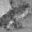} &
	\includegraphics[width=\plotw]{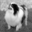} &
	\includegraphics[width=\plotw]{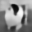} &
	\includegraphics[width=\plotw]{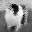} &
	\includegraphics[width=\plotw]{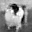} \\
	\includegraphics[width=\plotw]{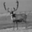} &
	\includegraphics[width=\plotw]{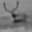} &
	\includegraphics[width=\plotw]{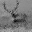} &
	\includegraphics[width=\plotw]{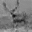} &
	\includegraphics[width=\plotw]{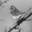} &
	\includegraphics[width=\plotw]{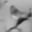} &
	\includegraphics[width=\plotw]{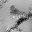} &
	\includegraphics[width=\plotw]{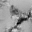} \\
	\includegraphics[width=\plotw]{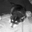} &
	\includegraphics[width=\plotw]{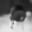} &
	\includegraphics[width=\plotw]{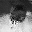} &
	\includegraphics[width=\plotw]{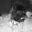} &
	\includegraphics[width=\plotw]{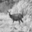} &
	\includegraphics[width=\plotw]{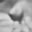} &
	\includegraphics[width=\plotw]{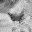} &
	\includegraphics[width=\plotw]{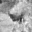} \\
	\includegraphics[width=\plotw]{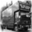} &
	\includegraphics[width=\plotw]{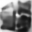} &
	\includegraphics[width=\plotw]{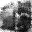} &
	\includegraphics[width=\plotw]{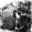} &
	\includegraphics[width=\plotw]{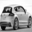} &
	\includegraphics[width=\plotw]{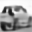} &
	\includegraphics[width=\plotw]{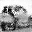} &
	\includegraphics[width=\plotw]{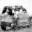} \\
	\includegraphics[width=\plotw]{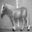} &
	\includegraphics[width=\plotw]{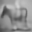} &
	\includegraphics[width=\plotw]{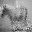} &
	\includegraphics[width=\plotw]{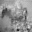} &
	\includegraphics[width=\plotw]{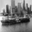} &
	\includegraphics[width=\plotw]{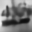} &
	\includegraphics[width=\plotw]{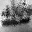} &
	\includegraphics[width=\plotw]{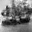} \\
	\includegraphics[width=\plotw]{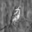} &
	\includegraphics[width=\plotw]{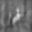} &
	\includegraphics[width=\plotw]{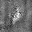} &
	\includegraphics[width=\plotw]{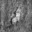} &
	\includegraphics[width=\plotw]{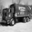} &
	\includegraphics[width=\plotw]{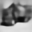} &
	\includegraphics[width=\plotw]{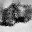} &
	\includegraphics[width=\plotw]{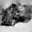} \\
	\includegraphics[width=\plotw]{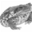} &
	\includegraphics[width=\plotw]{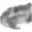} &
	\includegraphics[width=\plotw]{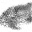} &
	\includegraphics[width=\plotw]{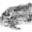} &
	\includegraphics[width=\plotw]{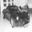} &
	\includegraphics[width=\plotw]{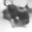} &
	\includegraphics[width=\plotw]{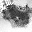} &
	\includegraphics[width=\plotw]{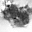} \\
	\includegraphics[width=\plotw]{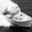} &
	\includegraphics[width=\plotw]{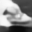} &
	\includegraphics[width=\plotw]{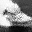} &
	\includegraphics[width=\plotw]{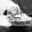} &
	\includegraphics[width=\plotw]{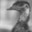} &
	\includegraphics[width=\plotw]{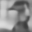} &
	\includegraphics[width=\plotw]{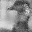} &
	\includegraphics[width=\plotw]{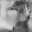} 
	\end{tabular}
	\caption{ Comparison of image reconstructions on the CIFAR10 dataset, where $\bmu$ is a reconstruction from a VAE, Diag. is $\bmu + \bepsilon$, where $\bepsilon$ is a sample from a diagonal covariance matrix, and in our method $\bepsilon$ is a sample from a dense covariance matrix.}
	\label{fig:cifar10_reconstructions}
\end{figure}

\newpage
\section{Splines dataset}
\label{sec:synthetic_datasets}

In the splines dataset, for each example, $\bmu$ is a cubic spline interpolation of five random points sampled from a Gaussian distribution, as shown in Fig.~\ref{subfig:splines_data_low_freq}.
Given a predefined prototype covariance matrix (shown in Fig.~\ref{subfig:splines_data_common_covar}),
the covariance matrix $\bSigma$ for a particular example is constructed by scaling this prototype covariance by the absolute value of $\bmu$, as shown in Fig.~\ref{subfig:splines_data_sample_covar}.
Finally, a random sample is drawn from the covariance matrix and added to the mean $\bmu$, which generates the final example $\bx = \bmu + \bepsilon$, as shown in Fig.~\ref{subfig:splines_data_low_plus_high_freq}.

\def\splDataW{0.25}

\begin{figure}[btph]
	\centering
	\setlength{\tabcolsep}{2pt}
	\begin{tabular}{cc}
	\begin{subfigure}{\splDataW\linewidth}
		\includegraphics[width=\linewidth]{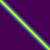}
		\caption{Prototype covariance}
		\label{subfig:splines_data_common_covar}
	\end{subfigure}
	&
	\begin{subfigure}{\splDataW\linewidth}
		\includegraphics[width=\linewidth]{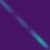}
		\caption{Example covariance}
		\label{subfig:splines_data_sample_covar}
	\end{subfigure}
	\\
	\begin{subfigure}{0.32\linewidth}
		\includegraphics[width=\linewidth]{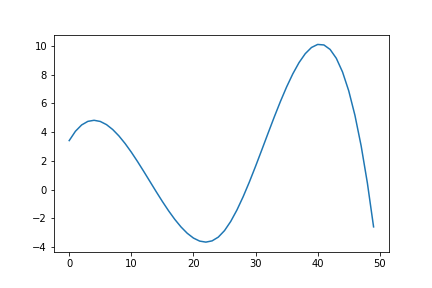}
		\caption{Example mean $\bmu$ \\ ~ \\} 
		\label{subfig:splines_data_low_freq}
	\end{subfigure}
	&
	\begin{subfigure}{0.32\linewidth}
		\includegraphics[width=\linewidth]{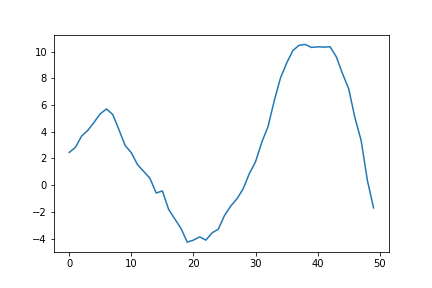}
		\caption{Example mean $\bmu$ plus sample from covariance}
		\label{subfig:splines_data_low_plus_high_freq}
	\end{subfigure}
	\end{tabular}\\[2pt]
	\caption{The splines dataset is synthesized using a prototype covariance matrix~(a), that is transformed to generate a number of example covariances~(b), where each transformation is function of a different spline~(c).
	Each example~(d) in the dataset is constructed by taking a single sample from a covariance (b) and adding it to the corresponding spline~(c).}
	\label{fig:splines_data}
\end{figure}

\newpage
\section{CelebA}
\label{sec:celeba}

\def\plotw{0.1\linewidth}

\def\1img{0}
\def\2img{9}
\def\3img{2}
\def\4img{11}
\def\5img{4}
\def\6img{5}
\def\7img{6}
\def\8img{7}
\def\9img{8}
\def\nEig{1000} 

\begin{figure}[htbp]
	\centering
	\setlength{\tabcolsep}{4pt} 
	\begin{tabular}{ccccccc}
	{\small Original } & {\small Input} & {\small DAE} & {\small Ours} &  {\small VAE-Recons} &
	{\small Difference} & {\small Proj. difference} \\
	& $\bx$ & & $\bmu + f(\bs)$ & $\bmu$ & $\bs = \bx - \bmu$ & $f(\bs)$ \\
	\includegraphics[width=\plotw]{supp_img/celeba/denoising2/input/\1img} &
	\includegraphics[width=\plotw]{supp_img/celeba/denoising2/input_noisy/\1img} &
	\includegraphics[width=\plotw]{supp_img/celeba/denoising2/denoising_ae/\1img} &
	\includegraphics[width=\plotw]{supp_img/celeba/denoising2/covar/denoised/\nEig/\1img} &
	\includegraphics[width=\plotw]{supp_img/celeba/denoising2/covar/recons/\1img} &
	\includegraphics[width=\plotw]{supp_img/celeba/denoising2/covar/residuals_noise/\1img} &
	\includegraphics[width=\plotw]{supp_img/celeba/denoising2/covar/proj_noise/\nEig/\1img} \\
	\includegraphics[width=\plotw]{supp_img/celeba/denoising2/input/\2img} &
	\includegraphics[width=\plotw]{supp_img/celeba/denoising2/input_noisy/\2img} &
	\includegraphics[width=\plotw]{supp_img/celeba/denoising2/denoising_ae/\2img} &
	\includegraphics[width=\plotw]{supp_img/celeba/denoising2/covar/denoised/\nEig/\2img} &
	\includegraphics[width=\plotw]{supp_img/celeba/denoising2/covar/recons/\2img} &
	\includegraphics[width=\plotw]{supp_img/celeba/denoising2/covar/residuals_noise/\2img} &
	\includegraphics[width=\plotw]{supp_img/celeba/denoising2/covar/proj_noise/\nEig/\2img} \\
	\includegraphics[width=\plotw]{supp_img/celeba/denoising2/input/\3img} &
	\includegraphics[width=\plotw]{supp_img/celeba/denoising2/input_noisy/\3img} &
	\includegraphics[width=\plotw]{supp_img/celeba/denoising2/denoising_ae/\3img} &
	\includegraphics[width=\plotw]{supp_img/celeba/denoising2/covar/denoised/\nEig/\3img} &
	\includegraphics[width=\plotw]{supp_img/celeba/denoising2/covar/recons/\3img} &
	\includegraphics[width=\plotw]{supp_img/celeba/denoising2/covar/residuals_noise/\3img} &
	\includegraphics[width=\plotw]{supp_img/celeba/denoising2/covar/proj_noise/\nEig/\3img} \\
	\includegraphics[width=\plotw]{supp_img/celeba/denoising2/input/\4img} &
	\includegraphics[width=\plotw]{supp_img/celeba/denoising2/input_noisy/\4img} &
	\includegraphics[width=\plotw]{supp_img/celeba/denoising2/denoising_ae/\4img} &
	\includegraphics[width=\plotw]{supp_img/celeba/denoising2/covar/denoised/\nEig/\4img} &
	\includegraphics[width=\plotw]{supp_img/celeba/denoising2/covar/recons/\4img} &
	\includegraphics[width=\plotw]{supp_img/celeba/denoising2/covar/residuals_noise/\4img} &
	\includegraphics[width=\plotw]{supp_img/celeba/denoising2/covar/proj_noise/\nEig/\4img} \\
	\includegraphics[width=\plotw]{supp_img/celeba/denoising2/input/\5img} &
	\includegraphics[width=\plotw]{supp_img/celeba/denoising2/input_noisy/\5img} &
	\includegraphics[width=\plotw]{supp_img/celeba/denoising2/denoising_ae/\5img} &
	\includegraphics[width=\plotw]{supp_img/celeba/denoising2/covar/denoised/\nEig/\5img} &
	\includegraphics[width=\plotw]{supp_img/celeba/denoising2/covar/recons/\5img} &
	\includegraphics[width=\plotw]{supp_img/celeba/denoising2/covar/residuals_noise/\5img} &
	\includegraphics[width=\plotw]{supp_img/celeba/denoising2/covar/proj_noise/\nEig/\5img} \\
	\includegraphics[width=\plotw]{supp_img/celeba/denoising2/input/\6img} &
	\includegraphics[width=\plotw]{supp_img/celeba/denoising2/input_noisy/\6img} &
	\includegraphics[width=\plotw]{supp_img/celeba/denoising2/denoising_ae/\6img} &
	\includegraphics[width=\plotw]{supp_img/celeba/denoising2/covar/denoised/\nEig/\6img} &
	\includegraphics[width=\plotw]{supp_img/celeba/denoising2/covar/recons/\6img} &
	\includegraphics[width=\plotw]{supp_img/celeba/denoising2/covar/residuals_noise/\6img} &
	\includegraphics[width=\plotw]{supp_img/celeba/denoising2/covar/proj_noise/\nEig/\6img} \\
	\includegraphics[width=\plotw]{supp_img/celeba/denoising2/input/\7img} &
	\includegraphics[width=\plotw]{supp_img/celeba/denoising2/input_noisy/\7img} &
	\includegraphics[width=\plotw]{supp_img/celeba/denoising2/denoising_ae/\7img} &
	\includegraphics[width=\plotw]{supp_img/celeba/denoising2/covar/denoised/\nEig/\7img} &
	\includegraphics[width=\plotw]{supp_img/celeba/denoising2/covar/recons/\7img} &
	\includegraphics[width=\plotw]{supp_img/celeba/denoising2/covar/residuals_noise/\7img} &
	\includegraphics[width=\plotw]{supp_img/celeba/denoising2/covar/proj_noise/\nEig/\7img} \\
	\includegraphics[width=\plotw]{supp_img/celeba/denoising2/input/\8img} &
	\includegraphics[width=\plotw]{supp_img/celeba/denoising2/input_noisy/\8img} &
	\includegraphics[width=\plotw]{supp_img/celeba/denoising2/denoising_ae/\8img} &
	\includegraphics[width=\plotw]{supp_img/celeba/denoising2/covar/denoised/\nEig/\8img} &
	\includegraphics[width=\plotw]{supp_img/celeba/denoising2/covar/recons/\8img} &
	\includegraphics[width=\plotw]{supp_img/celeba/denoising2/covar/residuals_noise/\8img} &
	\includegraphics[width=\plotw]{supp_img/celeba/denoising2/covar/proj_noise/\nEig/\8img} \\
	\includegraphics[width=\plotw]{supp_img/celeba/denoising2/input/\9img} &
	\includegraphics[width=\plotw]{supp_img/celeba/denoising2/input_noisy/\9img} &
	\includegraphics[width=\plotw]{supp_img/celeba/denoising2/denoising_ae/\9img} &
	\includegraphics[width=\plotw]{supp_img/celeba/denoising2/covar/denoised/\nEig/\9img} &
	\includegraphics[width=\plotw]{supp_img/celeba/denoising2/covar/recons/\9img} &
	\includegraphics[width=\plotw]{supp_img/celeba/denoising2/covar/residuals_noise/\9img} &
	\includegraphics[width=\plotw]{supp_img/celeba/denoising2/covar/proj_noise/\nEig/\9img} 
	\end{tabular}
	\caption{
	 Denoising experiment, left column: original image without noise, second column: image with added noise, third column: denoising autoencoder (DAE) result, fourth column: our result, fifth column: VAE reconstruction from the noisy input, sixth column: difference between the VAE reconstruction and the noisy input, right column: the difference projected on $\hat{\bSigma}$, the matrix constructed with 1000 eigenvectors of $\bSigma$ with the largest eigenvalues.
	Our result is the sum of the projected difference and the VAE reconstruction.
	Our model is able to recover fine details that are lost with the DAE approach.}
\end{figure}

\def\plotw{0.094\linewidth}

\begin{figure}[htbp]
	\centering
	\setlength{\tabcolsep}{1pt} 
	\begin{tabular}{ccccc @{\hspace{5\tabcolsep}} | @{\hspace{5\tabcolsep}} ccccc}
	{\small Input } & {\small AE} & {\small AE-Ours} & {\small VAE} & {\small VAE-Ours} & {\small Input } & {\small AE} & {\small AE-Ours} & {\small VAE} & {\small VAE-Ours}   \\
	\includegraphics[width=\plotw]{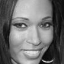} &
	\includegraphics[width=\plotw]{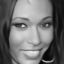} &
	\includegraphics[width=\plotw]{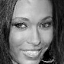} &
	\includegraphics[width=\plotw]{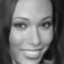} &
	\includegraphics[width=\plotw]{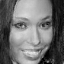} &
	\includegraphics[width=\plotw]{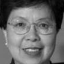} &
	\includegraphics[width=\plotw]{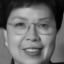} &
	\includegraphics[width=\plotw]{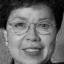} &
	\includegraphics[width=\plotw]{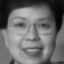} &
	\includegraphics[width=\plotw]{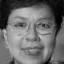}
	\\
	\includegraphics[width=\plotw]{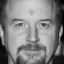} &
	\includegraphics[width=\plotw]{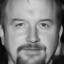} &
	\includegraphics[width=\plotw]{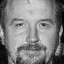} &
	\includegraphics[width=\plotw]{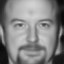} &
	\includegraphics[width=\plotw]{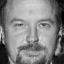} &
	\includegraphics[width=\plotw]{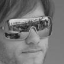} &
	\includegraphics[width=\plotw]{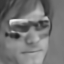} &
	\includegraphics[width=\plotw]{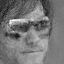} &
	\includegraphics[width=\plotw]{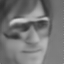} &
	\includegraphics[width=\plotw]{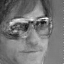}
	\\
	\includegraphics[width=\plotw]{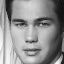} &
	\includegraphics[width=\plotw]{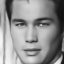} &
	\includegraphics[width=\plotw]{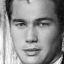} &
	\includegraphics[width=\plotw]{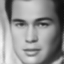} &
	\includegraphics[width=\plotw]{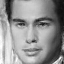} &
	\includegraphics[width=\plotw]{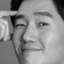} &
	\includegraphics[width=\plotw]{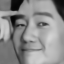} &
	\includegraphics[width=\plotw]{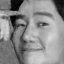} &
	\includegraphics[width=\plotw]{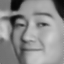} &
	\includegraphics[width=\plotw]{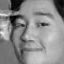}
	\\
	\includegraphics[width=\plotw]{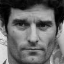} &
	\includegraphics[width=\plotw]{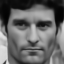} &
	\includegraphics[width=\plotw]{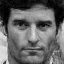} &
	\includegraphics[width=\plotw]{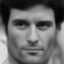} &
	\includegraphics[width=\plotw]{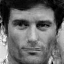} &
	\includegraphics[width=\plotw]{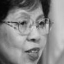} &
	\includegraphics[width=\plotw]{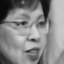} &
	\includegraphics[width=\plotw]{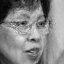} &
	\includegraphics[width=\plotw]{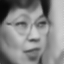} &
	\includegraphics[width=\plotw]{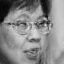}
	\\
	\includegraphics[width=\plotw]{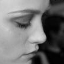} &
	\includegraphics[width=\plotw]{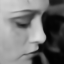} &
	\includegraphics[width=\plotw]{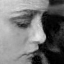} &
	\includegraphics[width=\plotw]{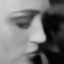} &
	\includegraphics[width=\plotw]{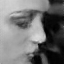} &
	\includegraphics[width=\plotw]{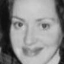} &
	\includegraphics[width=\plotw]{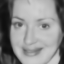} &
	\includegraphics[width=\plotw]{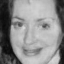} &
	\includegraphics[width=\plotw]{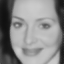} &
	\includegraphics[width=\plotw]{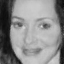}
	\\
	\includegraphics[width=\plotw]{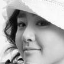} &
	\includegraphics[width=\plotw]{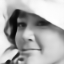} &
	\includegraphics[width=\plotw]{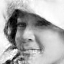} &
	\includegraphics[width=\plotw]{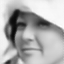} &
	\includegraphics[width=\plotw]{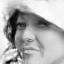} &
	\includegraphics[width=\plotw]{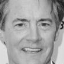} &
	\includegraphics[width=\plotw]{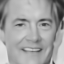} &
	\includegraphics[width=\plotw]{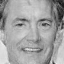} &
	\includegraphics[width=\plotw]{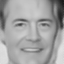} &
	\includegraphics[width=\plotw]{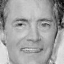}
	\\
	\includegraphics[width=\plotw]{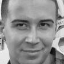} &
	\includegraphics[width=\plotw]{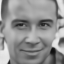} &
	\includegraphics[width=\plotw]{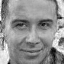} &
	\includegraphics[width=\plotw]{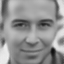} &
	\includegraphics[width=\plotw]{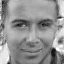} &
	\includegraphics[width=\plotw]{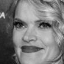} &
	\includegraphics[width=\plotw]{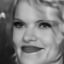} &
	\includegraphics[width=\plotw]{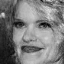} &
	\includegraphics[width=\plotw]{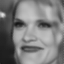} &
	\includegraphics[width=\plotw]{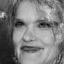}
	\\
	\includegraphics[width=\plotw]{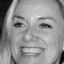} &
	\includegraphics[width=\plotw]{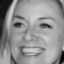} &
	\includegraphics[width=\plotw]{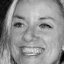} &
	\includegraphics[width=\plotw]{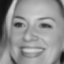} &
	\includegraphics[width=\plotw]{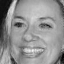} &
	\includegraphics[width=\plotw]{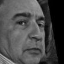} &
	\includegraphics[width=\plotw]{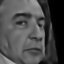} &
	\includegraphics[width=\plotw]{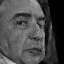} &
	\includegraphics[width=\plotw]{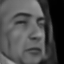} &
	\includegraphics[width=\plotw]{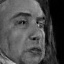}
	\end{tabular}
	\caption{Comparison of image reconstructions for the different models.
	The best of 100 and 1000 samples from $\bSigma$ measured using the MSE to the input are shown.
	The AE and VAE both generate over-smoothed images.
	For both the AE and VAE, our model adds plausible high-frequencies from a single sample drawn from the predicted uncertainty distribution.}
\end{figure}

\def\plotw{0.1\linewidth}

\begin{figure}[btph]
	\centering
	\setlength{\tabcolsep}{1pt} 
	\begin{tabular}{cc @{\hspace{5\tabcolsep}}  |  @{\hspace{5\tabcolsep}} cc  @{\hspace{5\tabcolsep}}  |  @{\hspace{5\tabcolsep}}  cc @{\hspace{5\tabcolsep}}  |  @{\hspace{5\tabcolsep}} cc}
	{\small $\bmu$ } & {\small Ours} & {\small $\bmu$ } & {\small Ours} & {\small $\bmu$ } & {\small Ours} & {\small $\bmu$ } & {\small Ours} \\
	\includegraphics[width=\plotw]{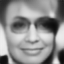} &
	\includegraphics[width=\plotw]{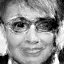} & 
	\includegraphics[width=\plotw]{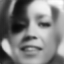} &
	\includegraphics[width=\plotw]{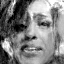} & 
	\includegraphics[width=\plotw]{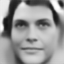} &
	\includegraphics[width=\plotw]{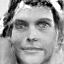} &
	\includegraphics[width=\plotw]{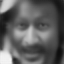} &
	\includegraphics[width=\plotw]{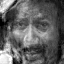} \\
	\includegraphics[width=\plotw]{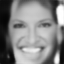} &
	\includegraphics[width=\plotw]{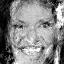} & 
	\includegraphics[width=\plotw]{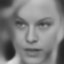} &
	\includegraphics[width=\plotw]{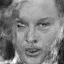} & 
	\includegraphics[width=\plotw]{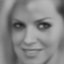} &
	\includegraphics[width=\plotw]{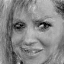} &
	\includegraphics[width=\plotw]{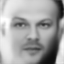} &
	\includegraphics[width=\plotw]{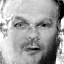} \\
	\includegraphics[width=\plotw]{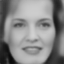} &
	\includegraphics[width=\plotw]{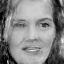} & 
	\includegraphics[width=\plotw]{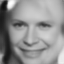} &
	\includegraphics[width=\plotw]{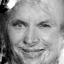} & 
	\includegraphics[width=\plotw]{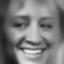} &
	\includegraphics[width=\plotw]{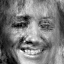} &
	\includegraphics[width=\plotw]{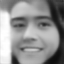} &
	\includegraphics[width=\plotw]{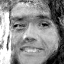} \\
	\includegraphics[width=\plotw]{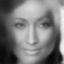} &
	\includegraphics[width=\plotw]{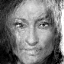} & 
	\includegraphics[width=\plotw]{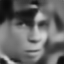} &
	\includegraphics[width=\plotw]{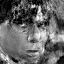} & 
	\includegraphics[width=\plotw]{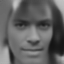} &
	\includegraphics[width=\plotw]{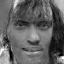} &
	\includegraphics[width=\plotw]{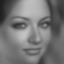} &
	\includegraphics[width=\plotw]{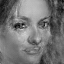} \\
	\includegraphics[width=\plotw]{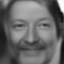} &
	\includegraphics[width=\plotw]{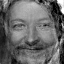} & 
	\includegraphics[width=\plotw]{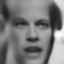} &
	\includegraphics[width=\plotw]{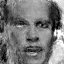} & 
	\includegraphics[width=\plotw]{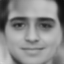} &
	\includegraphics[width=\plotw]{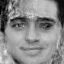} &
	\includegraphics[width=\plotw]{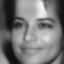} &
	\includegraphics[width=\plotw]{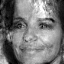} \\
	\includegraphics[width=\plotw]{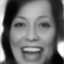} &
	\includegraphics[width=\plotw]{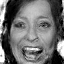} & 
	\includegraphics[width=\plotw]{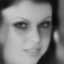} &
	\includegraphics[width=\plotw]{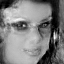} & 
	\includegraphics[width=\plotw]{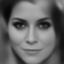} &
	\includegraphics[width=\plotw]{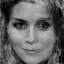} &
	\includegraphics[width=\plotw]{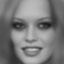} &
	\includegraphics[width=\plotw]{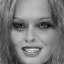} \\
	\includegraphics[width=\plotw]{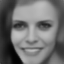} &
	\includegraphics[width=\plotw]{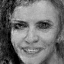} & 
	\includegraphics[width=\plotw]{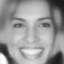} &
	\includegraphics[width=\plotw]{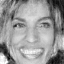} & 
	\includegraphics[width=\plotw]{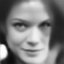} &
	\includegraphics[width=\plotw]{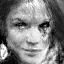} &
	\includegraphics[width=\plotw]{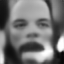} &
	\includegraphics[width=\plotw]{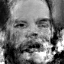} \\
	\includegraphics[width=\plotw]{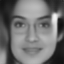} &
	\includegraphics[width=\plotw]{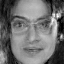} & 
	\includegraphics[width=\plotw]{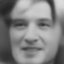} &
	\includegraphics[width=\plotw]{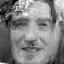} & 
	\includegraphics[width=\plotw]{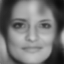} &
	\includegraphics[width=\plotw]{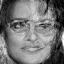} &
	\includegraphics[width=\plotw]{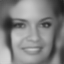} &
	\includegraphics[width=\plotw]{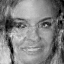} \\
	\includegraphics[width=\plotw]{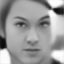} &
	\includegraphics[width=\plotw]{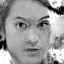} & 
	\includegraphics[width=\plotw]{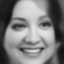} &
	\includegraphics[width=\plotw]{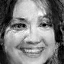} & 
	\includegraphics[width=\plotw]{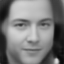} &
	\includegraphics[width=\plotw]{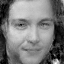} &
	\includegraphics[width=\plotw]{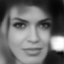} &
	\includegraphics[width=\plotw]{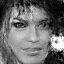} \\
	\includegraphics[width=\plotw]{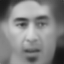} &
	\includegraphics[width=\plotw]{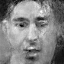} & 
	\includegraphics[width=\plotw]{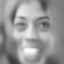} &
	\includegraphics[width=\plotw]{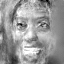} & 
	\includegraphics[width=\plotw]{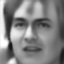} &
	\includegraphics[width=\plotw]{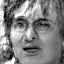} &
	\includegraphics[width=\plotw]{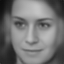} &
	\includegraphics[width=\plotw]{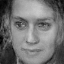}
	\end{tabular}
	\caption{Images generated by decoding samples from the prior distribution on the latent space of a $\beta$-VAE with added residuals from our model, where $\beta=5$.}
	\label{fig:celeba_sampleskl5}
\end{figure}

\def\plotw{0.1\linewidth}

\begin{figure}[btph]
	\centering
	\setlength{\tabcolsep}{1pt} 
	\begin{tabular}{cc @{\hspace{5\tabcolsep}}  |  @{\hspace{5\tabcolsep}} cc  @{\hspace{5\tabcolsep}}  |  @{\hspace{5\tabcolsep}}  cc @{\hspace{5\tabcolsep}}  |  @{\hspace{5\tabcolsep}} cc}
	{\small $\bmu$ } & {\small Ours} & {\small $\bmu$ } & {\small Ours} & {\small $\bmu$ } & {\small Ours} & {\small $\bmu$ } & {\small Ours} \\
	\includegraphics[width=\plotw]{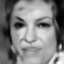} &
	\includegraphics[width=\plotw]{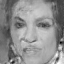} & 
	\includegraphics[width=\plotw]{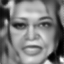} &
	\includegraphics[width=\plotw]{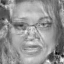} & 
	\includegraphics[width=\plotw]{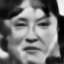} &
	\includegraphics[width=\plotw]{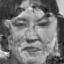} &
	\includegraphics[width=\plotw]{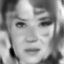} &
	\includegraphics[width=\plotw]{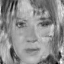} \\
	\includegraphics[width=\plotw]{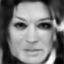} &
	\includegraphics[width=\plotw]{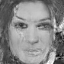} & 
	\includegraphics[width=\plotw]{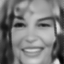} &
	\includegraphics[width=\plotw]{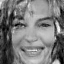} & 
	\includegraphics[width=\plotw]{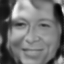} &
	\includegraphics[width=\plotw]{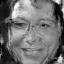} &
	\includegraphics[width=\plotw]{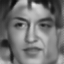} &
	\includegraphics[width=\plotw]{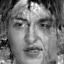} \\
	\includegraphics[width=\plotw]{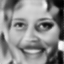} &
	\includegraphics[width=\plotw]{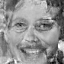} & 
	\includegraphics[width=\plotw]{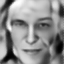} &
	\includegraphics[width=\plotw]{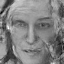} & 
	\includegraphics[width=\plotw]{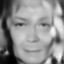} &
	\includegraphics[width=\plotw]{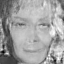} &
	\includegraphics[width=\plotw]{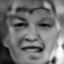} &
	\includegraphics[width=\plotw]{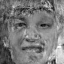} \\
	\includegraphics[width=\plotw]{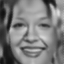} &
	\includegraphics[width=\plotw]{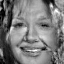} & 
	\includegraphics[width=\plotw]{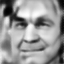} &
	\includegraphics[width=\plotw]{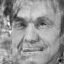} & 
	\includegraphics[width=\plotw]{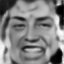} &
	\includegraphics[width=\plotw]{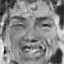} &
	\includegraphics[width=\plotw]{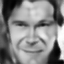} &
	\includegraphics[width=\plotw]{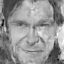} \\
	\includegraphics[width=\plotw]{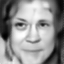} &
	\includegraphics[width=\plotw]{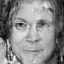} & 
	\includegraphics[width=\plotw]{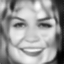} &
	\includegraphics[width=\plotw]{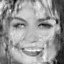} & 
	\includegraphics[width=\plotw]{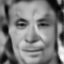} &
	\includegraphics[width=\plotw]{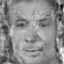} &
	\includegraphics[width=\plotw]{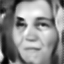} &
	\includegraphics[width=\plotw]{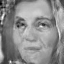} \\
	\includegraphics[width=\plotw]{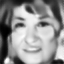} &
	\includegraphics[width=\plotw]{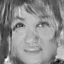} & 
	\includegraphics[width=\plotw]{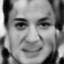} &
	\includegraphics[width=\plotw]{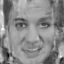} & 
	\includegraphics[width=\plotw]{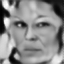} &
	\includegraphics[width=\plotw]{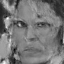} &
	\includegraphics[width=\plotw]{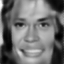} &
	\includegraphics[width=\plotw]{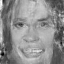} \\
	\includegraphics[width=\plotw]{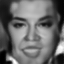} &
	\includegraphics[width=\plotw]{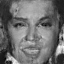} & 
	\includegraphics[width=\plotw]{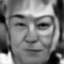} &
	\includegraphics[width=\plotw]{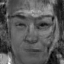} & 
	\includegraphics[width=\plotw]{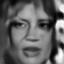} &
	\includegraphics[width=\plotw]{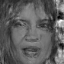} &
	\includegraphics[width=\plotw]{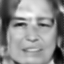} &
	\includegraphics[width=\plotw]{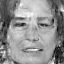} \\
	\includegraphics[width=\plotw]{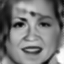} &
	\includegraphics[width=\plotw]{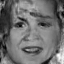} & 
	\includegraphics[width=\plotw]{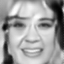} &
	\includegraphics[width=\plotw]{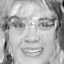} & 
	\includegraphics[width=\plotw]{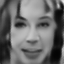} &
	\includegraphics[width=\plotw]{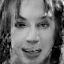} &
	\includegraphics[width=\plotw]{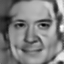} &
	\includegraphics[width=\plotw]{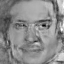}
	\end{tabular}
	\caption{Images generated by decoding samples from the prior distribution on the latent space of a VAE.}
\end{figure}

\def\plotw{0.11\linewidth}

\def\1img{0}
\def\2img{1}
\def\3img{2}
\def\4img{3}
\def\5img{4}
\def\6img{5}
\def\7img{6}
\def\8img{7}
\def\9img{8}
\def\a1img{9}

\begin{figure*}[btph]
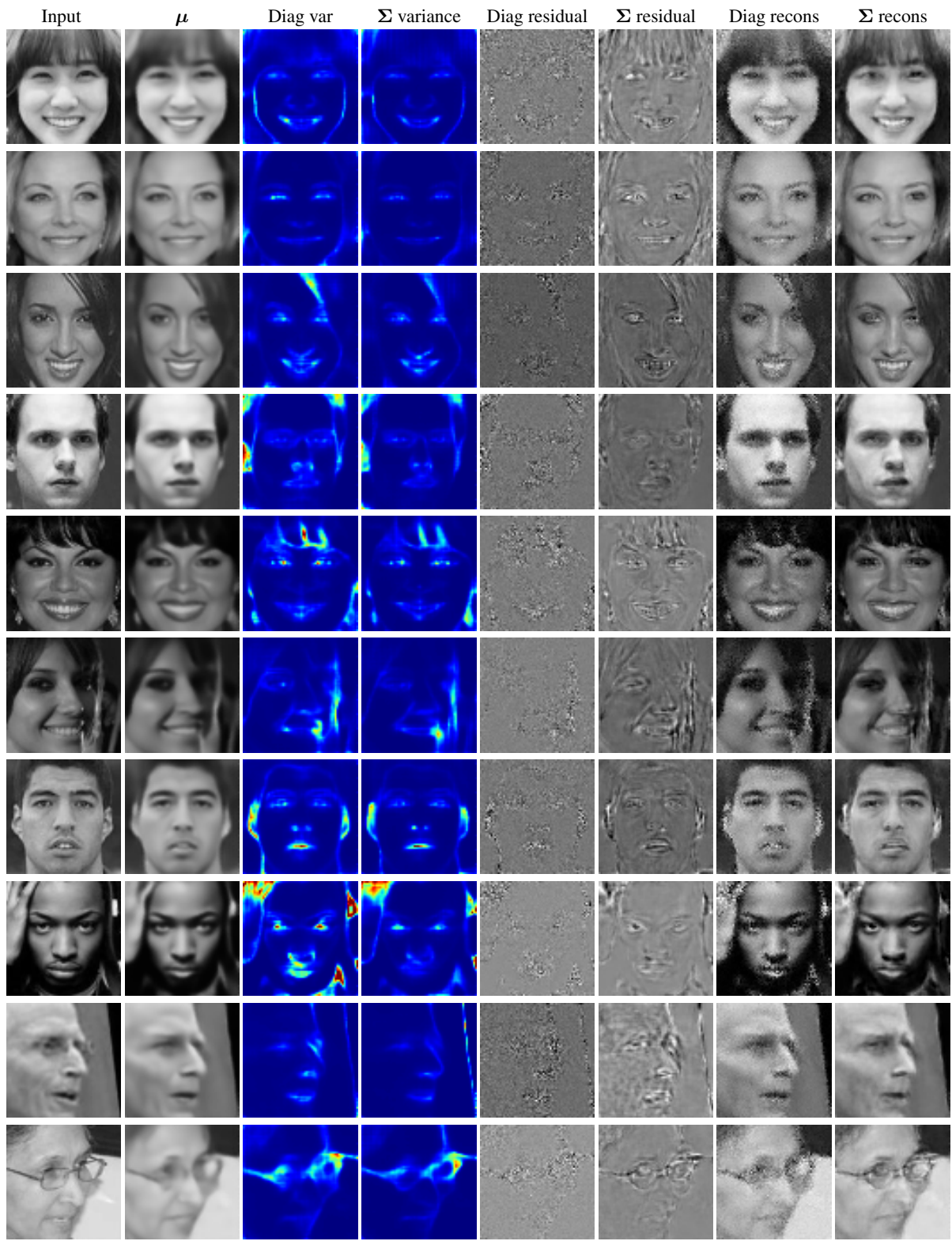

	\centering
	\setlength{\tabcolsep}{1pt} 
	\begin{tabular}{cccccccc}
	{\small Input } & {\small $\bmu$} & {\small Diag var} & {\small $\bSigma$ variance} & {\small Diag residual} & {\small $\bSigma$ residual} & {\small Diag recons} & {\small $\bSigma$ recons}\\
	\includegraphics[width=\plotw]{supp_img/celeba/variance_fig/input/\1img} &
	\includegraphics[width=\plotw]{supp_img/celeba/variance_fig/recons/\1img} &
	\includegraphics[width=\plotw]{supp_img/celeba/variance_fig/variance_diag/\1img} &
	\includegraphics[width=\plotw]{supp_img/celeba/variance_fig/variance_covar/\1img} &
	\includegraphics[width=\plotw]{supp_img/celeba/variance_fig/diag/residuals/\1img} &
	\includegraphics[width=\plotw]{supp_img/celeba/variance_fig/covar/residuals//\1img} &
	\includegraphics[width=\plotw]{supp_img/celeba/variance_fig/diag/recons_samples/\1img} &
	\includegraphics[width=\plotw]{supp_img/celeba/variance_fig/covar/recons_samples/\1img}
	\\
	\includegraphics[width=\plotw]{supp_img/celeba/variance_fig/input/\2img} &
	\includegraphics[width=\plotw]{supp_img/celeba/variance_fig/recons/\2img} &
	\includegraphics[width=\plotw]{supp_img/celeba/variance_fig/variance_diag/\2img} &
	\includegraphics[width=\plotw]{supp_img/celeba/variance_fig/variance_covar/\2img} &
	\includegraphics[width=\plotw]{supp_img/celeba/variance_fig/diag/residuals/\2img} &
	\includegraphics[width=\plotw]{supp_img/celeba/variance_fig/covar/residuals//\2img} &
	\includegraphics[width=\plotw]{supp_img/celeba/variance_fig/diag/recons_samples/\2img} &
	\includegraphics[width=\plotw]{supp_img/celeba/variance_fig/covar/recons_samples/\2img}
	\\
	\includegraphics[width=\plotw]{supp_img/celeba/variance_fig/input/\3img} &
	\includegraphics[width=\plotw]{supp_img/celeba/variance_fig/recons/\3img} &
	\includegraphics[width=\plotw]{supp_img/celeba/variance_fig/variance_diag/\3img} &
	\includegraphics[width=\plotw]{supp_img/celeba/variance_fig/variance_covar/\3img} &
	\includegraphics[width=\plotw]{supp_img/celeba/variance_fig/diag/residuals/\3img} &
	\includegraphics[width=\plotw]{supp_img/celeba/variance_fig/covar/residuals//\3img} &
	\includegraphics[width=\plotw]{supp_img/celeba/variance_fig/diag/recons_samples/\3img} &
	\includegraphics[width=\plotw]{supp_img/celeba/variance_fig/covar/recons_samples/\3img}
	\\
	\includegraphics[width=\plotw]{supp_img/celeba/variance_fig/input/\4img} &
	\includegraphics[width=\plotw]{supp_img/celeba/variance_fig/recons/\4img} &
	\includegraphics[width=\plotw]{supp_img/celeba/variance_fig/variance_diag/\4img} &
	\includegraphics[width=\plotw]{supp_img/celeba/variance_fig/variance_covar/\4img} &
	\includegraphics[width=\plotw]{supp_img/celeba/variance_fig/diag/residuals/\4img} &
	\includegraphics[width=\plotw]{supp_img/celeba/variance_fig/covar/residuals//\4img} &
	\includegraphics[width=\plotw]{supp_img/celeba/variance_fig/diag/recons_samples/\4img} &
	\includegraphics[width=\plotw]{supp_img/celeba/variance_fig/covar/recons_samples/\4img}
	\\
	\includegraphics[width=\plotw]{supp_img/celeba/variance_fig/input/\5img} &
	\includegraphics[width=\plotw]{supp_img/celeba/variance_fig/recons/\5img} &
	\includegraphics[width=\plotw]{supp_img/celeba/variance_fig/variance_diag/\5img} &
	\includegraphics[width=\plotw]{supp_img/celeba/variance_fig/variance_covar/\5img} &
	\includegraphics[width=\plotw]{supp_img/celeba/variance_fig/diag/residuals/\5img} &
	\includegraphics[width=\plotw]{supp_img/celeba/variance_fig/covar/residuals//\5img} &
	\includegraphics[width=\plotw]{supp_img/celeba/variance_fig/diag/recons_samples/\5img} &
	\includegraphics[width=\plotw]{supp_img/celeba/variance_fig/covar/recons_samples/\5img}
	\\
	\includegraphics[width=\plotw]{supp_img/celeba/variance_fig/input/\6img} &
	\includegraphics[width=\plotw]{supp_img/celeba/variance_fig/recons/\6img} &
	\includegraphics[width=\plotw]{supp_img/celeba/variance_fig/variance_diag/\6img} &
	\includegraphics[width=\plotw]{supp_img/celeba/variance_fig/variance_covar/\6img} &
	\includegraphics[width=\plotw]{supp_img/celeba/variance_fig/diag/residuals/\6img} &
	\includegraphics[width=\plotw]{supp_img/celeba/variance_fig/covar/residuals//\6img} &
	\includegraphics[width=\plotw]{supp_img/celeba/variance_fig/diag/recons_samples/\6img} &
	\includegraphics[width=\plotw]{supp_img/celeba/variance_fig/covar/recons_samples/\6img}
	\\
	\includegraphics[width=\plotw]{supp_img/celeba/variance_fig/input/\7img} &
	\includegraphics[width=\plotw]{supp_img/celeba/variance_fig/recons/\7img} &
	\includegraphics[width=\plotw]{supp_img/celeba/variance_fig/variance_diag/\7img} &
	\includegraphics[width=\plotw]{supp_img/celeba/variance_fig/variance_covar/\7img} &
	\includegraphics[width=\plotw]{supp_img/celeba/variance_fig/diag/residuals/\7img} &
	\includegraphics[width=\plotw]{supp_img/celeba/variance_fig/covar/residuals//\7img} &
	\includegraphics[width=\plotw]{supp_img/celeba/variance_fig/diag/recons_samples/\7img} &
	\includegraphics[width=\plotw]{supp_img/celeba/variance_fig/covar/recons_samples/\7img}
	\\
	\includegraphics[width=\plotw]{supp_img/celeba/variance_fig/input/\8img} &
	\includegraphics[width=\plotw]{supp_img/celeba/variance_fig/recons/\8img} &
	\includegraphics[width=\plotw]{supp_img/celeba/variance_fig/variance_diag/\8img} &
	\includegraphics[width=\plotw]{supp_img/celeba/variance_fig/variance_covar/\8img} &
	\includegraphics[width=\plotw]{supp_img/celeba/variance_fig/diag/residuals/\8img} &
	\includegraphics[width=\plotw]{supp_img/celeba/variance_fig/covar/residuals//\8img} &
	\includegraphics[width=\plotw]{supp_img/celeba/variance_fig/diag/recons_samples/\8img} &
	\includegraphics[width=\plotw]{supp_img/celeba/variance_fig/covar/recons_samples/\8img}
	\\
	\includegraphics[width=\plotw]{supp_img/celeba/variance_fig/input/\9img} &
	\includegraphics[width=\plotw]{supp_img/celeba/variance_fig/recons/\9img} &
	\includegraphics[width=\plotw]{supp_img/celeba/variance_fig/variance_diag/\9img} &
	\includegraphics[width=\plotw]{supp_img/celeba/variance_fig/variance_covar/\9img} &
	\includegraphics[width=\plotw]{supp_img/celeba/variance_fig/diag/residuals/\9img} &
	\includegraphics[width=\plotw]{supp_img/celeba/variance_fig/covar/residuals//\9img} &
	\includegraphics[width=\plotw]{supp_img/celeba/variance_fig/diag/recons_samples/\9img} &
	\includegraphics[width=\plotw]{supp_img/celeba/variance_fig/covar/recons_samples/\9img}
	\\
	\includegraphics[width=\plotw]{supp_img/celeba/variance_fig/input/\a1img} &
	\includegraphics[width=\plotw]{supp_img/celeba/variance_fig/recons/\a1img} &
	\includegraphics[width=\plotw]{supp_img/celeba/variance_fig/variance_diag/\a1img} &
	\includegraphics[width=\plotw]{supp_img/celeba/variance_fig/variance_covar/\a1img} &
	\includegraphics[width=\plotw]{supp_img/celeba/variance_fig/diag/residuals/\a1img} &
	\includegraphics[width=\plotw]{supp_img/celeba/variance_fig/covar/residuals//\a1img} &
	\includegraphics[width=\plotw]{supp_img/celeba/variance_fig/diag/recons_samples/\a1img} &
	\includegraphics[width=\plotw]{supp_img/celeba/variance_fig/covar/recons_samples/\a1img}
	\end{tabular}
	\caption{ Variance maps for different inputs, taking a single sample from the estimated residual distribution.
	The diagonal noise estimation model mistakenly identifies teeth or skin wrinkles as variance, whereas the covariance model properly identifies them as regions with high covariance, yet low variance.}
\end{figure*}

\def\plotw{0.11\linewidth}

\def\1img{2}
\def\2img{6}
\def\3img{2}
\def\4img{7}
\def\6img{0}

\begin{figure}[t!]
	\centering
	\setlength{\tabcolsep}{2pt} 
	\begin{tabular}{c @{\hspace{5pt}} cccccc @{\hspace{5pt}} c}
	{\small Source} & & & & & & & {\small Target} \\
	\multirow{2}{*}{\includegraphics[align=c, width=\plotw]{supp_img/celeba/interp_fig/input_source/\6img/0}} &
	\includegraphics[align=c, width=\plotw]{supp_img/celeba/interp_fig/recons_with/\6img/0} &
	\includegraphics[align=c, width=\plotw]{supp_img/celeba/interp_fig/recons_with/\6img/2} &
	\includegraphics[align=c, width=\plotw]{supp_img/celeba/interp_fig/recons_with/\6img/4} &
	\includegraphics[align=c, width=\plotw]{supp_img/celeba/interp_fig/recons_with/\6img/6} &
	\includegraphics[align=c, width=\plotw]{supp_img/celeba/interp_fig/recons_with/\6img/8} & 
	\includegraphics[align=c, width=\plotw]{supp_img/celeba/interp_fig/recons_with/\6img/10} &
	\multirow{2}{*}{\includegraphics[align=c, width=\plotw]{supp_img/celeba/interp_fig/input_target/\6img/0}}
	\\[25pt]
	 & 
	\includegraphics[align=c, width=\plotw]{supp_img/celeba/interp_fig/recons/\6img/0} &
	\includegraphics[align=c, width=\plotw]{supp_img/celeba/interp_fig/recons/\6img/2} &
	\includegraphics[align=c, width=\plotw]{supp_img/celeba/interp_fig/recons/\6img/4} &
	\includegraphics[align=c, width=\plotw]{supp_img/celeba/interp_fig/recons/\6img/6} &
	\includegraphics[align=c, width=\plotw]{supp_img/celeba/interp_fig/recons/\6img/8} & 
	\includegraphics[align=c, width=\plotw]{supp_img/celeba/interp_fig/recons/\6img/10} &
	\\[25pt]
	\multirow{2}{*}{\includegraphics[align=c, width=\plotw]{supp_img/celeba/interp_fig/input_source/\3img/0}} &
	\includegraphics[align=c, width=\plotw]{supp_img/celeba/interp_fig/recons_with/\3img/0} &
	\includegraphics[align=c, width=\plotw]{supp_img/celeba/interp_fig/recons_with/\3img/2} &
	\includegraphics[align=c, width=\plotw]{supp_img/celeba/interp_fig/recons_with/\3img/4} &
	\includegraphics[align=c, width=\plotw]{supp_img/celeba/interp_fig/recons_with/\3img/6} &
	\includegraphics[align=c, width=\plotw]{supp_img/celeba/interp_fig/recons_with/\3img/8} & 
	\includegraphics[align=c, width=\plotw]{supp_img/celeba/interp_fig/recons_with/\3img/10} &
	\multirow{2}{*}{\includegraphics[align=c, width=\plotw]{supp_img/celeba/interp_fig/input_target/\3img/0}}
	\\[25pt]
	 &
	\includegraphics[align=c, width=\plotw]{supp_img/celeba/interp_fig/recons/\3img/0} &
	\includegraphics[align=c, width=\plotw]{supp_img/celeba/interp_fig/recons/\3img/2} &
	\includegraphics[align=c, width=\plotw]{supp_img/celeba/interp_fig/recons/\3img/4} &
	\includegraphics[align=c, width=\plotw]{supp_img/celeba/interp_fig/recons/\3img/6} &
	\includegraphics[align=c, width=\plotw]{supp_img/celeba/interp_fig/recons/\3img/8} & 
	\includegraphics[align=c, width=\plotw]{supp_img/celeba/interp_fig/recons/\3img/10} &
	\\[25pt]
	\multirow{2}{*}{\includegraphics[align=c, width=\plotw]{supp_img/celeba/interp_fig_2/input_source/\1img/0}} &
	\includegraphics[align=c, width=\plotw]{supp_img/celeba/interp_fig_2/recons_with/\1img/0} &
	\includegraphics[align=c, width=\plotw]{supp_img/celeba/interp_fig_2/recons_with/\1img/2} &
	\includegraphics[align=c, width=\plotw]{supp_img/celeba/interp_fig_2/recons_with/\1img/4} &
	\includegraphics[align=c, width=\plotw]{supp_img/celeba/interp_fig_2/recons_with/\1img/6} &
	\includegraphics[align=c, width=\plotw]{supp_img/celeba/interp_fig_2/recons_with/\1img/8} & 
	\includegraphics[align=c, width=\plotw]{supp_img/celeba/interp_fig_2/recons_with/\1img/10} &
	\multirow{2}{*}{\includegraphics[align=c, width=\plotw]{supp_img/celeba/interp_fig_2/input_target/\1img/0}}
	\\[25pt]
	&
	\includegraphics[align=c, width=\plotw]{supp_img/celeba/interp_fig_2/recons/\1img/0} &
	\includegraphics[align=c, width=\plotw]{supp_img/celeba/interp_fig_2/recons/\1img/2} &
	\includegraphics[align=c, width=\plotw]{supp_img/celeba/interp_fig_2/recons/\1img/4} &
	\includegraphics[align=c, width=\plotw]{supp_img/celeba/interp_fig_2/recons/\1img/6} &
	\includegraphics[align=c, width=\plotw]{supp_img/celeba/interp_fig_2/recons/\1img/8} & 
	\includegraphics[align=c, width=\plotw]{supp_img/celeba/interp_fig_2/recons/\1img/10} &
	\\[25pt]
	\multirow{2}{*}{\includegraphics[align=c, width=\plotw]{supp_img/celeba/interp_fig_2/input_source/\2img/0}} &
	\includegraphics[align=c, width=\plotw]{supp_img/celeba/interp_fig_2/recons_with/\2img/0} &
	\includegraphics[align=c, width=\plotw]{supp_img/celeba/interp_fig_2/recons_with/\2img/2} &
	\includegraphics[align=c, width=\plotw]{supp_img/celeba/interp_fig_2/recons_with/\2img/4} &
	\includegraphics[align=c, width=\plotw]{supp_img/celeba/interp_fig_2/recons_with/\2img/6} &
	\includegraphics[align=c, width=\plotw]{supp_img/celeba/interp_fig_2/recons_with/\2img/8} & 
	\includegraphics[align=c, width=\plotw]{supp_img/celeba/interp_fig_2/recons_with/\2img/10} &
	\multirow{2}{*}{\includegraphics[align=c, width=\plotw]{supp_img/celeba/interp_fig_2/input_target/\2img/0}}
	\\[25pt]
	&
	\includegraphics[align=c, width=\plotw]{supp_img/celeba/interp_fig_2/recons/\2img/0} &
	\includegraphics[align=c, width=\plotw]{supp_img/celeba/interp_fig_2/recons/\2img/2} &
	\includegraphics[align=c, width=\plotw]{supp_img/celeba/interp_fig_2/recons/\2img/4} &
	\includegraphics[align=c, width=\plotw]{supp_img/celeba/interp_fig_2/recons/\2img/6} &
	\includegraphics[align=c, width=\plotw]{supp_img/celeba/interp_fig_2/recons/\2img/8} & 
	\includegraphics[align=c, width=\plotw]{supp_img/celeba/interp_fig_2/recons/\2img/10} &
	\\[25pt]
	\multirow{2}{*}{\includegraphics[align=c, width=\plotw]{supp_img/celeba/interp_fig/input_source/\4img/0}} &
	\includegraphics[align=c, width=\plotw]{supp_img/celeba/interp_fig/recons_with/\4img/0} &
	\includegraphics[align=c, width=\plotw]{supp_img/celeba/interp_fig/recons_with/\4img/2} &
	\includegraphics[align=c, width=\plotw]{supp_img/celeba/interp_fig/recons_with/\4img/4} &
	\includegraphics[align=c, width=\plotw]{supp_img/celeba/interp_fig/recons_with/\4img/6} &
	\includegraphics[align=c, width=\plotw]{supp_img/celeba/interp_fig/recons_with/\4img/8} & 
	\includegraphics[align=c, width=\plotw]{supp_img/celeba/interp_fig/recons_with/\4img/10} &
	\multirow{2}{*}{\includegraphics[align=c, width=\plotw]{supp_img/celeba/interp_fig/input_target/\4img/0}}
	\\[25pt]
	&
	\includegraphics[align=c, width=\plotw]{supp_img/celeba/interp_fig/recons/\4img/0} &
	\includegraphics[align=c, width=\plotw]{supp_img/celeba/interp_fig/recons/\4img/2} &
	\includegraphics[align=c, width=\plotw]{supp_img/celeba/interp_fig/recons/\4img/4} &
	\includegraphics[align=c, width=\plotw]{supp_img/celeba/interp_fig/recons/\4img/6} &
	\includegraphics[align=c, width=\plotw]{supp_img/celeba/interp_fig/recons/\4img/8} & 
	\includegraphics[align=c, width=\plotw]{supp_img/celeba/interp_fig/recons/\4img/10} &
	\end{tabular}
	\caption{ Samples drawn with our model while interpolating on the latent space, from the source to target.
	Using a fixed noise vector $\mathbf{u}$, a single sample is drawn from our model as $\bx = \bmu + \bM \mathbf{u}$, where the covariance $\bSigma = \bM\bM^T$ is estimated from each $\bz$.
	The latent values $\bz$ are interpolated using polar interpolation.
	For every pair of rows, the first contains the result from our model, and the second contains the reconstruction $\bmuz$ from the VAE.}
\end{figure}

\end{document}